\documentclass{article}

\PassOptionsToPackage{numbers}{natbib}

\usepackage[preprint]{neurips_2026}
  
\usepackage[canadian]{babel}
\usepackage[utf8]{inputenc} 
\usepackage[T1]{fontenc}    
\usepackage{hyperref}      
\usepackage{url}            
\usepackage{booktabs}       
\usepackage{amsfonts}       
\usepackage{nicefrac}      
\usepackage{microtype}      
\usepackage{xcolor}
\usepackage{comment}
\usepackage{graphicx}
\usepackage{upgreek}
\usepackage{gensymb}
\usepackage{pgfplots}
\usepackage{pgfplotstable}
\pgfplotsset{compat=1.18}
\usepgfplotslibrary{statistics}
\usepgfplotslibrary{groupplots}   
\usepackage{booktabs}
\usepackage{multirow}
\usepackage{pgf-pie}
\usepackage{subcaption}
\usepackage{tikz}
\usepackage{multirow}
\usepackage{graphicx}
\usepackage{tabularx}
\usepackage{array}
\usepackage{siunitx}
\usepackage{tikz}
\usepackage{xcolor}
\usepackage{hyperref}
\usepackage{subcaption}
\newcolumntype{C}[1]{>{\centering\arraybackslash}m{#1}}
\definecolor{phaseA}{RGB}{31,119,180}   
\definecolor{phaseB}{RGB}{255,127,14}   
\definecolor{phaseC}{RGB}{44,160,44}    
\definecolor{phaseD}{RGB}{214,39,40}   
\definecolor{phaseE}{RGB}{148,103,189}

\usepackage{pifont}
\usepackage[acronym]{glossaries}
\usepackage{glossaries-prefix}
\usepackage{wrapfig}
\usepackage{float}
\usepackage{listings}
\usepackage{xcolor}

\definecolor{vsKey}{RGB}{156, 220, 254}      
\definecolor{vsKeyDark}{RGB}{30, 116, 185}   
\definecolor{vsString}{RGB}{206, 145, 120}   
\definecolor{vsStringDark}{RGB}{163, 21, 21} 
\definecolor{vsNumber}{RGB}{181, 206, 168}   
\definecolor{vsNumberDark}{RGB}{9, 134, 88}  
\definecolor{vsPunct}{RGB}{60, 60, 60}       
\definecolor{vsBg}{RGB}{250, 250, 250}       

\lstdefinelanguage{json}{
    basicstyle=\ttfamily\footnotesize\color{vsPunct},
    backgroundcolor=\color{vsBg},
    showstringspaces=false,
    breaklines=true,
    frame=single,
    framesep=4pt,
    framerule=0.4pt,
    rulecolor=\color{gray!40},
    string=[s]{"}{"},
    stringstyle=\color{vsStringDark},
    morestring=[b]",
    literate=
        *{0}{{{\color{vsNumberDark}0}}}{1}
         {1}{{{\color{vsNumberDark}1}}}{1}
         {2}{{{\color{vsNumberDark}2}}}{1}
         {3}{{{\color{vsNumberDark}3}}}{1}
         {4}{{{\color{vsNumberDark}4}}}{1}
         {5}{{{\color{vsNumberDark}5}}}{1}
         {6}{{{\color{vsNumberDark}6}}}{1}
         {7}{{{\color{vsNumberDark}7}}}{1}
         {8}{{{\color{vsNumberDark}8}}}{1}
         {9}{{{\color{vsNumberDark}9}}}{1}
}
\definecolor{ForestGreen}{RGB}{34,139,34}

\title{REBOOT: From Failure to Recovery – A Dataset and Benchmark for Precision Assembly}

\author{%
  Nana Yaw Owusu Ofori-Ampofo \\
  University of Calgary\\
  \texttt{nana.oforiampofo@ucalgary.ca} \\
  \And
  Samira Ebrahimi Kahou \\
  University of Calgary, Mila \\
  \texttt{samira.ebrahimikahou@ucalgary.ca} \\
  \AND
  Joseph Thekinen \\
  University of Calgary \\
  \texttt{joseph.thekinen@ucalgary.ca} \\
}

\begin{document}

\maketitle

\begin{abstract}
    Robot learning policies fail in characterizable ways: they stall at high-uncertainty states, drift during contact-rich alignment, and miss targets by millimetres on precision tasks. Yet training datasets consist almost exclusively of successful demonstrations, and real-world benchmarks typically collapse performance into binary success. This disconnect leaves policies without supervision for recovery behaviours and leaves researchers without the resolution needed to localize and analyze failure over long horizons. 
    We introduce REBOOT (Recovery Episode Benchmark for Off-nominal Trajectories), the first robot manipulation benchmark designed around failure as a first-class signal. REBOOT comprises 2,160 demonstrations across 18 precision assembly tasks, each decomposed into a shared five-phase sequence — Align(pick) → Engage(pick) → Transport → Align(place) → Engage(place) — enabling per-phase progress evaluation beyond terminal success. Failures are systematically introduced at each phase and paired with expert recovery trajectories that return the system to a valid continuation state. Each task is annotated with structured metadata encoding mechanical sources of difficulty: mating interfaces are labelled by rotational symmetry class (continuous, C2, or C1), geometric precision tier based on engagement clearance, and assembly direction via matched install–remove pairs. Failure episodes are annotated by phase of occurrence and categorical failure mode (e.g., misalignment, jamming, premature release), enabling fine-grained attribution of failure to kinematic phase and tolerance violation. All data are collected with synchronized RGB-D observations from four fixed camera viewpoints. Each phase is further accompanied by grounded natural-language outcome descriptions characterizing both success and failure conditions (e.g., stalling in free space, misaligned grasp, insertion jamming), providing supervision for vision-language recovery policies. Half the dataset consists of expert demonstrations; the other half consists of recovery-from-failure demonstrations sampled to match the empirical failure distribution of imitation-learned policy rollouts. We benchmark action-chunked transformer, diffusion, and $\pi_0$-FAST policies using per-phase completion rates, revealing architecture-specific failure points invisible under binary evaluation. Datasets and code are available at \url{https://nanayawoa.github.io/REBOOT}.
\end{abstract}

\section{Introduction}
Imitation learning on human teleportation data has become the dominant paradigm in robot manipulation, with recent academic and industrial efforts betting on massive data scaling as a path to generalist robot policies~\cite{embodimentcollaboration2025openxembodimentroboticlearning, khazatsky2024droid, walke2023bridgedata}. Yet results increasingly suggest this approach is hitting a performance ceiling well below reliable task completion. Even with thousands of demonstrations, state-of-the-art models plateau at 70--80\% success on seemingly simple tasks, and performance gains from additional data exhibit diminishing returns. Demonstrations are systematically biased toward clean, successful trajectories. But real deployment demands behaviours for handling compounding errors, stochasticity, and recovery from partial failure, which are behaviours absent from success-only training data.

\begin{figure}[t]
\centering
\includegraphics[width=\textwidth]{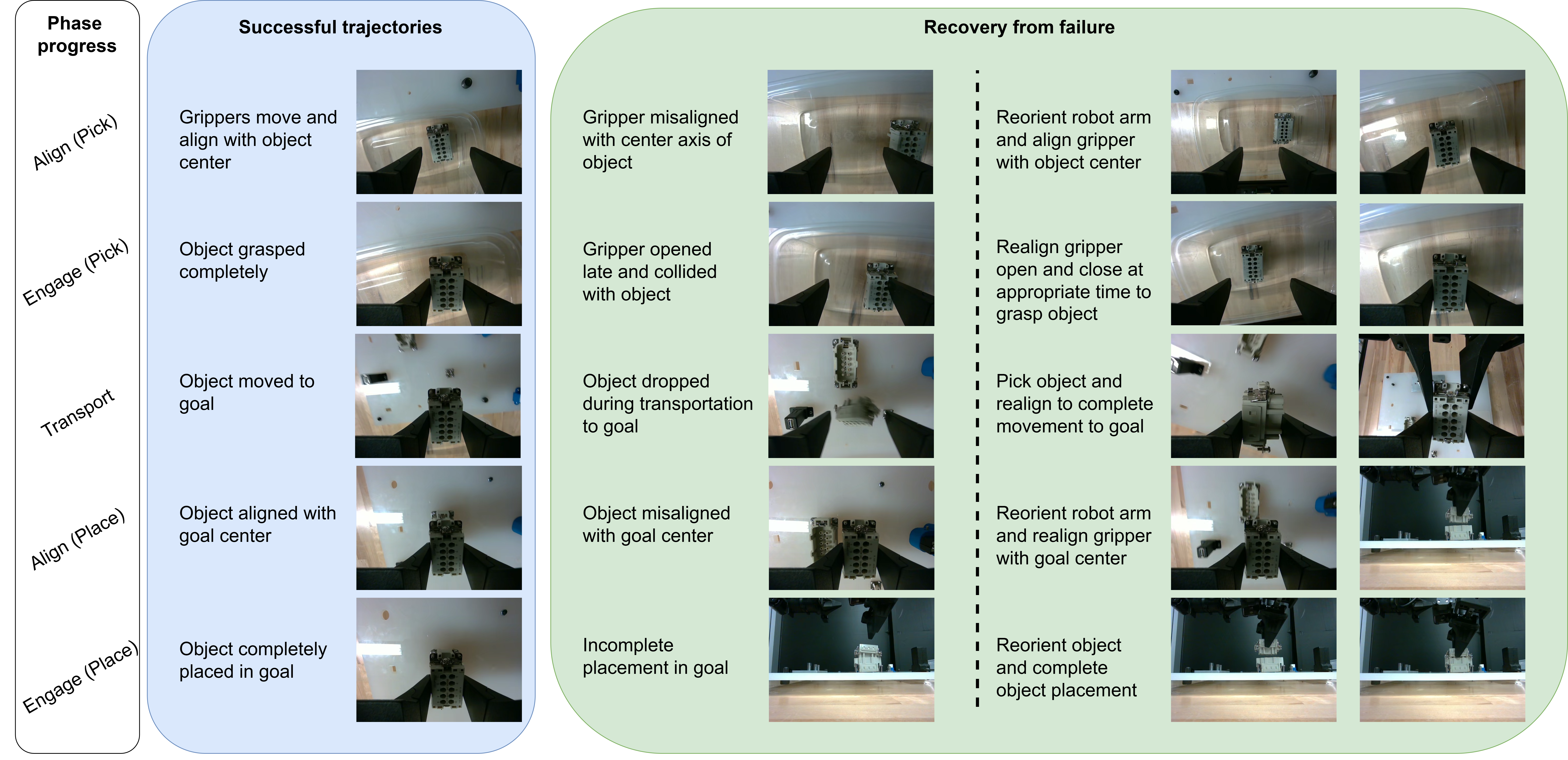}
\caption{Overview of REBOOT. Each precision-assembly task is decomposed into five kinematic phases---\textit{Align(pick)}, \textit{Engage(pick)}, \textit{Transport}, \textit{Align(place)}, and \textit{Engage(place)}---shown vertically on the left. The first column illustrates successful task execution with natural language annotations at each phase. The second column shows some failure modes that arise at each phase during autonomous policy rollouts, paired with the expert recovery trajectories that return the system to a valid continuation state. REBOOT treats both columns as first-class supervision signal.}
\label{fig:overview}
\end{figure}

This limitation is especially stark in precision assembly. Consider inserting a USB-A connector (0.62\,mm of engagement clearance) requiring a sequence of kinematic phases: grasp, transport, align, and mate. A policy trained only on successes has no exposure to mid-task deviation: arriving at alignment with a lateral offset exceeding the port clearance, it drives forward rather than retracting. The housing catches on the port rim, yaws under contact force, and jams irreversibly. What the policy needed, but was never shown, is a recovery trajectory: retract, re-align, re-approach. As shown in Figure~\ref{fig:overview}, the same kinematic phase produces qualitatively different outcomes depending on whether the policy has seen failure and recovery there.

As summarized in Table~\ref{tab:failure-datasets}, existing datasets either contain minimal failure data or lack structured recovery annotations. Recent efforts to close this gap through human-in-the-loop correction~\cite{hu2025racrobotlearninglonghorizon, luo2025precisedexterousroboticmanipulation}, targeted data augmentation~\cite{dai2024racerrichlanguageguidedfailure}, and failure-annotated datasets~\cite{lu2025robofac} represent meaningful progress, but share a common limitation: failures are organized by semantic category (wrong object, grasp miss) rather than by the kinematic phase in which they occur or the mechanical tolerance that was violated. This coarse vocabulary treats a sub-millimetre alignment drift and a multi-millimetre one as identical events, and provides no signal about where in a long-horizon task the policy began to diverge---resolution that is essential when tight clearances convert small errors into unrecoverable jamming.

\begin{table*}[t]
\centering
\small
\setlength{\tabcolsep}{4pt}
\renewcommand{\arraystretch}{1.0}
\caption{Robot manipulation datasets reporting explicit failure or recovery episode counts. Counts are taken verbatim from the primary source (paper text and dataset cards). REBOOT is the only real-robot dataset with a 50\% failure rate and structured per-phase recovery annotations.}
\label{tab:failure-datasets}
\begin{tabular}{lccccc}
\toprule
\textbf{Dataset} & \textbf{Domain} & \textbf{\# of demos} & \textbf{\% Failure} & \textbf{Recovery - Yes/No}  \\
\midrule
REFLECT\cite{liu2023reflect}     & Sim+Real & 130       & 100.0\% & No  \\
RoboFAC \cite{lu2025robofac}                 & Sim+Real & 10{,}722  & 88.0\%  & No  \\
RaC \cite{hu2025racrobotlearninglonghorizon} & Sim+Real & N/A$^{\dagger}$ & N/A$^{\dagger}$ & \textbf{Yes} \\
DROID \cite{khazatsky2024droid}              & Real     & $\approx$92{,}000 & $\approx$17.4\% & Yes (3.68\%)  \\
REASSEMBLE \cite{Sliwowski-RSS-25}    & Real     & 4{,}551   & 11.3\%  & No   \\
FurnitureBench \cite{heo2023furniturebench}  & Sim+Real & 5{,}100   & 0\%     & No  \\
RH20T \cite{fang2024rh20t}                   & Real     & $\approx$110{,}000 & $\approx$9.1\% & No  \\
RoboMIND \cite{wu2025robomind}               & Real     & 112{,}000 & 4.5\%   & No   \\
\midrule
\textbf{Ours}                                & Real     & 2{,}160   & 50\%    & \textbf{Yes} \\
\bottomrule
\end{tabular}
\par\smallskip
\footnotesize $^{\dagger}$RaC reports data scale in hours of teleoperation (5h for shirt-hanging) and recovery-to-correction frame ratios (1:1 to 1:2), not as a fixed demonstration count.
\end{table*}

We introduce \textbf{REBOOT} (Recovery Episode Benchmark for Off-nominal Trajectories), the first precision manipulation benchmark designed around failure as a first-class signal in the precision assembly regime, with a standardized phase decomposition and failure vocabulary that makes cross-task recovery analysis tractable for the first time. REBOOT contributes the following:
\begin{enumerate}

    \item \textbf{Open multi-sensor manipulation benchmark with paired failure data.}
    2,160 real-robot trajectories across 18 precision connector mating tasks~\cite{NIST2022}, spanning 0 to 2.73\,mm engagement clearance: 1,080 expert demonstrations paired with 1,080 recovery episodes whose per-phase failure distribution is calibrated to match autonomous policy rollouts. Each recovery episode is annotated with the phase of failure induction and a categorical failure mode (misalignment, jamming, premature release, off-axis collision).

    \item \textbf{A shared phase decomposition for assembly and disassembly.}
    A common five-phase sequence---\textit{Align(pick)} $\rightarrow$ \textit{Engage(pick)} $\rightarrow$ \textit{Transport} $\rightarrow$ \textit{Align(place)} $\rightarrow$ \textit{Engage(place)}---with annotated phase boundaries across all 18 tasks. This unified structure enables cross-task failure analysis and recovery transfer for the first time.

    \item \textbf{Structured mechanical metadata for failure attribution.}
    Each task is annotated with rotational symmetry class, engagement clearance tier, and matched install--remove pairs.

    \item \textbf{Grounded natural-language outcome descriptions.}
    Per-phase natural-language annotations characterizing success and failure conditions provide supervision for vision-language recovery policies.

    \item \textbf{Empirical failure distributions across three policy families.}
    We systematically record at which phase and how precision-assembly policies fail in practice for various policy classes and use that to calibrate success-recovery proportions in the dataset.

\end{enumerate}

We train and rollout three policy families Action Chunking with Transformers~\cite{zhao2023learningfinegrainedbimanualmanipulation}, Diffusion Policy~\cite{chi2024diffusionpolicy}, and $\pi_0$-FAST~\cite{intelligence2025pi05visionlanguageactionmodelopenworld} to calibrate the dataset. We systematically record the phase and failure mode at which policies fail in practice across all 18 tasks, to set the expert-to-recovery ratio in REBOOT rather than assuming a uniform mix. Benchmarking these three policy families using per-phase completion rates reveals architecture-specific failure concentrations in contact-rich engagement phases that are invisible under binary evaluation. REBOOT dataset is publicly released at \url{https://huggingface.co/REBOOT26/datasets}.

\section{Related work}

\paragraph{Benchmarks and evaluation for robotic assembly.}
Robotic manipulation benchmarks span simulation and real-world settings. Simulation benchmarks such as RLBench \cite{james2020rlbench} and ManiSkill2 \cite{gu2023maniskill2} provide broad task suites but face a sim-to-real gap on contact-rich tasks. Real-world benchmarks such as FurnitureBench \cite{heo2023furniturebench} and RoboEval \cite{wang2025roboevalroboticmanipulationmeets} evaluate real-world physical execution but operate at relatively coarse tolerances. The NIST Task Boards \cite{NIST2022} and REASSEMBLE \cite{Sliwowski-RSS-25} are the closest precedents, targeting precision assembly tasks such as connector mating.

Large-scale datasets such as Open X-Embodiment \cite{embodimentcollaboration2025openxembodimentroboticlearning}, DROID \cite{khazatsky2024droid}, and BridgeData V2 \cite{walke2023bridgedata} provide substantial coverage but consist almost exclusively of successful demonstrations. As summarized in Table~\ref{tab:failure-datasets}, existing datasets either contain minimal failure data or lack structured recovery annotations. Across both datasets and benchmarks, evaluation is typically reduced to binary success, obscuring differences between early failure and near-successful execution. REBOOT addresses this gap through a phase-decomposed benchmark with paired recovery trajectories, enabling fine-grained analysis of where and how failures occur within the task.

\paragraph{Learning from failure.}
Understanding \emph{how} learned policies fail, not just whether they succeed, has become a field of research in its own right in robotics and reinforcement learning. Empirical work shows that imitation-learned policies exhibit consistent failure patterns. Specific failure patterns reappear across tasks, architectures, and training regimes. For example, imitation-learned policies mirror subtle pauses in their demonstrations and stall at high-uncertainty states that call for decisive action \cite{chen2025exploitingpolicyidlingdexterous}. 

Two approaches have emerged in response: datasets that catalogue failures as supervision targets, and methods that synthesize them as a training signal. These approaches, although useful for advancing research in failure recovery, do not address the mechanical precision regime that dominates assembly tasks.

Several recent datasets treat failure annotation as a first-class component of manipulation benchmarks. RoboFAC \cite{lu2025robofac} releases a corpus covering six failure categories across three error levels, paired with a multi-modal critic. RoboMIND\cite{wu2025robomind} pairs 107k successful trajectories with 5k annotated failures across four embodiments. Neither targets the mechanical precision regime: failures are organized by semantic category (wrong object, grasp miss) rather than by the kinematic phase in which they occur or the tolerance that was violated. VLM-based failure reasoning\cite{duan2024aha} inherit the same coarse vocabulary, treating a \SI{0.1}{\milli\meter} alignment drift and a \SI{2}{\milli\meter} one as the same event. A complementary line probes rather than catalogues: RoboFail~\cite{sagar2026robomd} searches for environmental perturbations that break pre-trained policies, and STAR-Gen~\cite{gao2026taxonomy} organizes generalization failures along visual, semantic, and behavioural axes. Both characterize failures under \emph{distributional shift}, whereas precision assembly is dominated by \emph{in-distribution} failures — the robot attempts the task it was trained on exactly, but lacks the spatial precision it demands. RaC~\cite{hu2025racrobotlearninglonghorizon}, RACER~\cite{dai2024racerrichlanguageguidedfailure}, Guardian \cite{pacaud2026guardiandetectingroboticplanning} synthesize recovery or procedural failure variants to improve downstream policies. 

The shared limitation lies in which failures get represented: synthesis-focused methods over-produce modes that are easy to simulate (slippage and large pose offsets) and underproduce sub-millimetre misalignment and premature phase transitions that dominate real policy roll-outs on precision assembly tasks. Runtime detectors such as FAIL-Detect~\cite{xu2025detectfailuresfailuredata} sit on the output side, monitoring trained policies; they complement rather than compete with our input-side contribution.

These methods all \emph{consume} failure data; our work is designed to \emph{produce} it. Each task has recovery demonstrations recorded to represent the failure distribution measured on that task, and these episodes are annotated with the kinematic phase at which failure was induced, yielding a precision-regime resource that synthesis-based and semantically-categorized datasets do not provide.

\section{REBOOT dataset}

\subsection{Definitions}
\paragraph{Observation.}
At each time step $t$, an observation bundles synchronized sensor readings from the bi-manual platform:
\begin{equation}
o_t = \big(\mathbf{I}_t^{\mathrm{rgb}},\ \mathbf{I}_t^{\mathrm{depth}}\ s_t\big),
\end{equation}
where $\mathbf{I}_t^{\mathrm{rgb}} \in \mathbb{R}^{4 \times 480 \times 640 \times 3}$ and $\mathbf{I}_t^{\mathrm{depth}} \in \mathbb{R}^{4 \times 480 \times 640 \times 1}$ represent RGB and depth frames from four calibrated cameras (overhead, low-angle, and one per wrist), and $s_t = (q_t, g_t) \in \mathbb{R}^{14}$ concatenates per-arm joint positions $q_t \in \mathbb{R}^{12}$ (6 joints per arm) with the gripper state $g_t \in \mathbb{R}^{2}$. 

\paragraph{Action.}
At each time step $t$, the action specifies the target configuration commanded to the low-level controller:
\begin{equation}
a_t = \big(q_t^{\mathrm{cmd}},\ g_t^{\mathrm{cmd}}\big) \in \mathbb{R}^{14},
\end{equation}
where $q_t^{\mathrm{cmd}} \in \mathbb{R}^{12}$ are absolute target joint positions for both arms (6 joints per arm) and $g_t^{\mathrm{cmd}} \in \mathbb{R}^{2}$ are continuous target gripper positions for each end-effector. Actions are recorded at \SI{30}{\hertz} via teleoperation from the leader arms.

\paragraph{Trajectory.}
A time-aligned sequence of $T$ observation--action pairs:
\begin{equation}
    \mathcal{T} = \{(o_t,\, a_t)\}_{t=1}^{T}.
\end{equation}

\paragraph{Demonstration.}
A trajectory $\mathcal{T}^{\mathrm{demo}}$ collected via tele-operation, in which a human operator performs the task from start to completion. Multiple demonstrations are recorded per task to form the training corpus $\mathcal{D} = \{\mathcal{T}^{\mathrm{demo}}_i\}_{i=1}^{N}$ used to fit imitation learning policies.

\paragraph{Policy rollout.}
A trajectory $\mathcal{T}^{\mathrm{roll}}$ generated autonomously by a trained policy $\pi_\theta$ executing the task without human intervention, where actions are sampled as $a_t \sim \pi_\theta(\cdot \mid o_t)$.

\paragraph{Phase.}
\label{app:phase_def}
A contiguous sub-sequence of a trajectory corresponding to a semantically distinct stage of the assembly task. Each trajectory is partitioned into five ordered phases by annotated time-step boundaries $1 = \tau_0 < \tau_1 < \cdots < \tau_5 = T$, where the $k$-th phase spans $[\tau_{k-1}, \tau_k)$: (1)~\textit{Align(pick)}, positioning both arms to the grasp configuration; (2)~\textit{Engage(pick)}, closing the grippers to secure the object; (3)~\textit{Transport}, moving the grasped object to the target location; (4)~\textit{Align(place)}, orienting the object for placement; and (5)~\textit{Engage(place)}, releasing and seating the object at the goal pose. This five-phase decomposition is shared uniformly across all tasks. Each trajectory is annotated with the time-step boundaries between consecutive phases.

\paragraph{Recovery episode.}
A tele-operated demonstration that intentionally induces a common failure state observed during autonomous policy rollouts before recovering to successful task completion. Unlike standard demonstrations, recovery episodes are imperfect by design, exposing the policy to realistic failure modes. Each recovery episode is annotated with the phase index $k^* \in \{1,\dots,5\}$ at which the failure was induced, enabling phase-targeted recovery training.

\subsection{Hardware and recording setup}

Our dataset was collected using a WidowX AI stationary bi-manual tele-operation system. Dataset recording, policy training, and policy rollouts are executed through Trossen AI's fork of the open-source LeRobot framework\cite{cadene2024lerobot}, ensuring compatibility and enabling plug-and-play extension of our dataset to other researchers on LeRobot's software ecosystem. The WidowX family of robots is widely used in the robot learning community; it has been used in popular robot learning works such as BridgeData V2~\cite{walke2023bridgedata}, ALOHA~\cite{zhao2023learningfinegrainedbimanualmanipulation} and Mobile ALOHA~\cite{fu2024mobile}.

\subsection{Task design}
Our task suite is built on the well-established NIST Assembly Task Board~\cite{NIST2022}, specifically Task Board \#1, which represents a cross-section of precision assembly and disassembly tasks. The suite comprises 9 objects spanning a wide range of geometric complexity, engagement tolerances, and rotational symmetries, with each object appearing in both an install and a remove variant, yielding 18 tasks in total. Each task is annotated with the following structured metadata:

\paragraph{Rotational symmetry group.}
The symmetry of the mating interface governs the difficulty of the rotational alignment problem. The symmetry is labelled under three groups: \textit{Continuous}, where any in-plane orientation is valid (e.g.\ 16\,mm cylinder, RCA connector); \textit{C2}, where exactly two orientations satisfy the mating constraint (e.g.\ USB-C connector, M12 fastener); and \textit{C1}, where a unique orientation is required (e.g.\ NEMA 1-15P power plug, USB-A connector). This label characterizes the angular search space that the policy must resolve at engagement.

\begin{figure}[h]
    \centering
    \includegraphics[width=\textwidth]{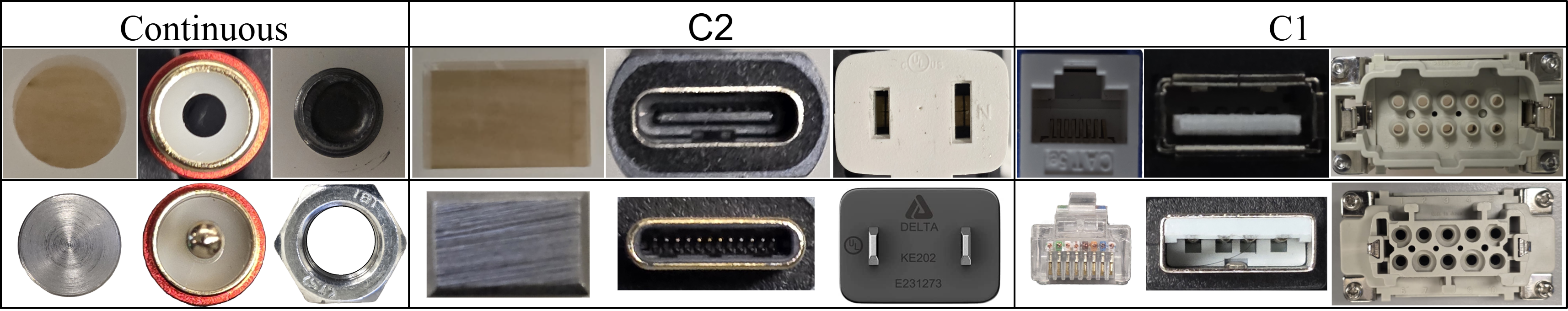}
    \caption{Rotational symmetry groups for representative objects in the dataset. Each column shows a goal--object pair; from left to right: 16\,mm cylinder (\textit{Continuous}), RCA connector (\textit{Continuous}), M12 fastener (\textit{C2}), 16\,mm bar (\textit{C2}), USB-C connector (\textit{C2}), NEMA 1-15P power plug (\textit{C1}), RJ45 connector (\textit{C1}), USB-A connector (\textit{C1}), and Harting HAN10E connector (\textit{C1}).}
\end{figure}

\paragraph{Phase labels.}
Each task is decomposed into the shared five-phase sequence defined in Section~\ref{app:phase_def}: \textit{Align(pick)} $\rightarrow$ \textit{Engage(pick)} $\rightarrow$ \textit{Transport} $\rightarrow$ \textit{Align(place)} $\rightarrow$ \textit{Engage(place)}. Phase transition timestamps are manually annotated for each trajectory. 

\paragraph{Geometric precision tier.}
Tasks are stratified into precision tiers according to their engagement tolerance. Tighter tolerances reduce the margin for misalignment, making successful engagement increasingly sensitive to small errors in perception and control, and thus serve as a direct proxy for task difficulty.

\subsection{Task phase decomposition}
Each task is decomposed into a shared sequence of five phases: \textit{Align(pick)} $\rightarrow$ \textit{Engage(pick)} $\rightarrow$ \textit{Transport} $\rightarrow$ \textit{Align(place)} $\rightarrow$ \textit{Engage(place)}. This task-agnostic decomposition enables cross-task analysis of where failures concentrate and supports direct comparison of per-phase success rates across tolerance tiers and symmetry groups. Beyond diagnostics, phase structure naturally supports goal-conditioned hierarchical policies that plan at the sub-task level before executing low-level actions. Hierarchical vision-language-action models such as $\pi_{0.5}$~\cite{intelligence2025pi05visionlanguageactionmodelopenworld}, which condition motor actions on predicted language sub-goals such as \textit{``pick up the cutting board''}.

\paragraph{Failure mode annotation.}
We annotate each failure episode with two attributes: the phase in which the failure occurred and the failure mode within that phase. The phase locates the failure in the trajectory; the mode identifies the mechanical event responsible. We define four failure modes shared across all tasks:
\begin{itemize}\itemsep2pt
    \item \textbf{Misalignment:} the end-effector pose deviates from the target alignment axis beyond the connector's clearance tolerance, preventing engagement.
    \item \textbf{Premature release:} the gripper opens before the object is fully seated or transferred.
    \item \textbf{Jamming:} the gripper or held object strikes the mating port off-axis during insertion, with contact forces spiking before alignment is achieved.
    \item \textbf{Off-axis collision:} the arm or held object collides with surrounding fixtures during transport or retraction, displacing the workpiece or triggering a safety stop.
\end{itemize}

\paragraph{Interaction with rotational symmetry.}
The frequency of each failure mode depends on the geometric structure of the connector. Connectors with \textit{Continuous} symmetry permit any rotation about the insertion axis, making rotational misalignment irrelevant and leaving translational misalignment as the dominant failure mode. Connectors with \textit{C2} symmetry introduce a discrete rotational constraint, making both translational and rotational misalignment distinct failure sub-modes. Connectors with \textit{C1} symmetry require simultaneous correct yaw (rotation about the insertion axis) and translational alignment along all three axes, and exhibit the highest misalignment failure rates in our data.

\paragraph{Interaction with clearance.}
Mechanical clearance sets the difficulty of the \textit{Align(place)} and \textit{Engage(place)} phases. Interference-fit connectors such as the RCA connector tolerate no positional error and convert any misalignment directly into jamming. Connectors with small clearances such as USB-C tolerate sub-millimeter error but amplify the gap between policy precision and demonstrated precision.  Loose connectors, such as the HAN 10E and NEMA 1-15P power plugs, tolerate larger positional errors, shifting the dominant failure mode from misalignment to off-axis collision and slip. The interaction of clearance with rotational symmetry produces the per-task progress profiles observed in Figure~\ref{fig:per-phase-failures}: tight, low-symmetry connectors fail earliest in the trajectory, while loose, high-symmetry connectors progress further before failing.

\subsection{Data collection protocol}

For each task, we collect 60 expert demonstrations and 60 recovery-from-failure episodes.

\paragraph{Expert demonstrations} The operator tele-operates the robot through the full task from start to completion. To introduce controlled positional diversity, the initial position of the task object is varied within a 3-6\,cm radius of a nominal start pose between episodes, while the goal position on the task board is held fixed. The 3-6\,cm range was chosen empirically: large enough to prevent policies from memorizing a fixed trajectory, small enough that 60 demonstrations cover it without requiring generalization across the full reachable workspace.

\paragraph{Recovery-from-failure demonstrations}Recovery episodes follow a protocol inspired by RaC~\cite{hu2025racrobotlearninglonghorizon} and are grounded in empirically observed failures rather than hypothesized ones. We first log the phase and failure mode of each failure that arises when policies trained on our expert demonstrations are run autonomously. The operator then teleoperates a recovery episode that reproduces the failure at its target phase (e.g. misalignment at Align(place)) and demonstrates the correction. Recording begins at object initialization, so each episode captures the full approach, the failure, and the recovery through to successful completion.\newline

\begin{figure}[H]
\centering
\includegraphics[width=\textwidth]{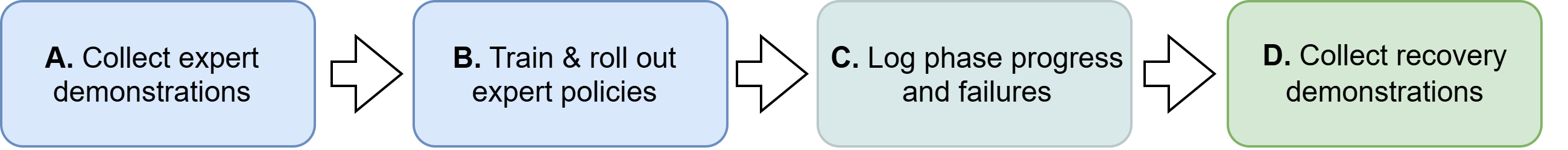}
\caption{Overview of the expert and recovery data collection pipeline. Policies trained on expert demonstrations are rolled out to identify phase-level failures, which are subsequently re-enacted by the teleoperator to collect successful recovery demonstrations.}
\label{fig:flowchart}
\end{figure}

\section{Dataset Statistics}
The full dataset comprises 2160 episodes across 18 tasks. Episode durations are standardized to 15, 30 and 60 seconds, depending on task complexity. Simple tasks such as USB cable removal typically complete within 10 seconds, with the remaining time as standby; complex tasks such as nut fastening utilize nearly the full 60-second window. Each episode contains 14-dimensional bi-manual joint position vectors (6 joints and 1 gripper state per arm) recorded at 30~Hz, four synchronized RGB-D video streams, and natural language annotations. The ratio of expert to failure-recovery episodes is 1:1 across all tasks, ensuring balanced coverage of nominal and failure-recovery behaviour.

\begin{wraptable}{r}{0.55\textwidth}
\centering
\begin{tabular}{lr}
\toprule
Connector & Clearance (mm) \\
\midrule
RCA           & interference fit (0) \\
USB-C         & 0.09 \\
RJ45          & 0.13 \\
16mm Bar      & 0.20 \\
16mm Cylinder & 0.20 \\
NEMA 1-15P power plug & 0.5  \\
USB-A         & 0.62 \\
M12 fastener   & 0.7 \\
HAN 10E       & 2.73 \\
\bottomrule
\end{tabular}
\caption{Connector clearance values.}
\label{tab:connector-clearance}
\end{wraptable}

\paragraph{Tasks and precision tiers.}
The 18 tasks span 10 connector and fastener types, organized into four precision tiers based on insertion clearance: \emph{interference fit}~(RCA: 0~mm clearance), \emph{tight}~(USB-C, RJ45: 0.09--0.13~mm), \emph{moderate}~(16~mm bar/cylinder,2-prong cable, USB-A: 0.20--0.62~mm), and \emph{loose}~(HAN~10E: 2.73~mm). Each task includes both an assembly variant and its disassembly counterpart, exposing both insertion-direction and extraction-direction failure modes within the same physical setup. We analyze and annotate our data along three dimensions: phase progress, failure modes, and recovery calibration.

\paragraph{Phase progress and failure anntation} We annotate each episode with timestamps marking the five phase boundaries as defined in \ref{app:phase_def}. Each failure episode is annotated with the phase at which the failure occurred and a categorical failure mode within that phase (misalignment, slip, premature release, jamming, or off-axis collision). 

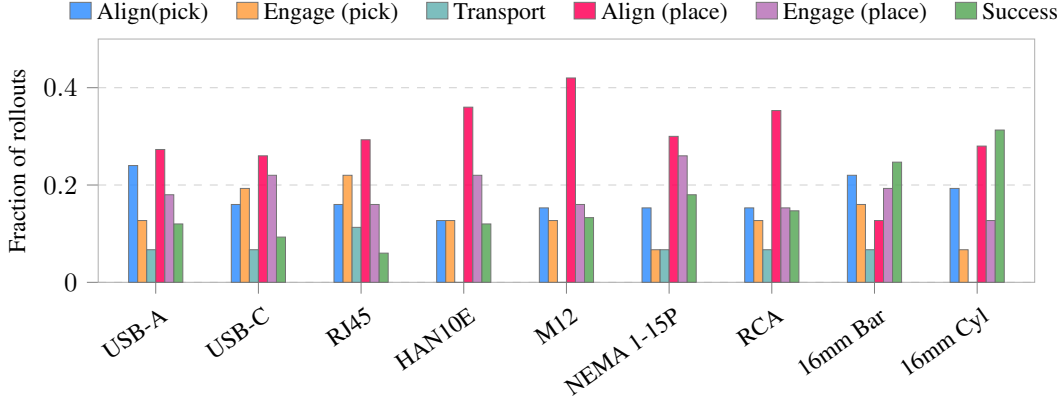
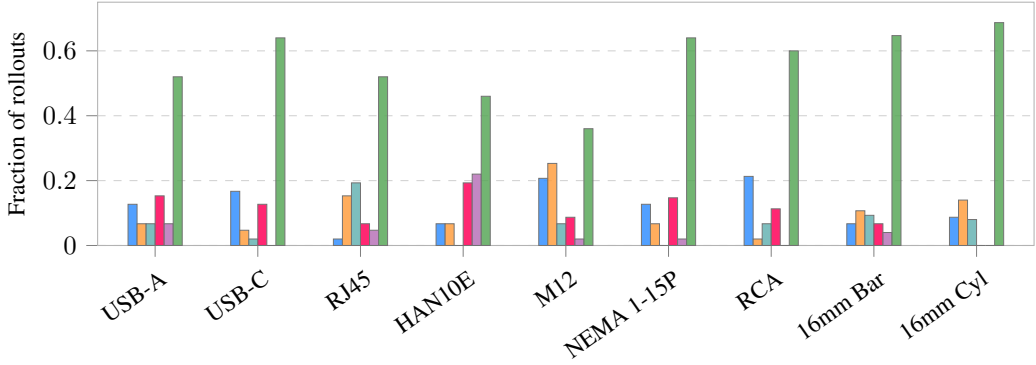
\begin{figure}[H]
\centering
\begin{subfigure}{\textwidth}
\centering
\begin{tikzpicture}
\begin{axis}[
    width=\textwidth,
    height=4.8cm,
    ybar=0pt,
    bar width=3.4pt,
    ymin=0, ymax=0.5,
    ytick={0,0.2,0.4,0.6},
    ylabel={Fraction of rollouts},
    ylabel style={font=\small},
    symbolic x coords={USB-A, USB-C, RJ45, HAN10E, M12, NEMA 1-15P,
                       RCA, 16mm Bar, 16mm Cyl},
    xtick=data,
    x tick label style={
        font=\small,
        rotate=35,
        anchor=north east,
        yshift=-1pt,
        xshift=2pt,
    },
    enlarge x limits=0.07,
    tick align=outside,
    tick pos=left,
    ymajorgrids,
    major grid style={dashed, gray!35, line width=0.4pt},
    axis line style={gray!60},
    legend style={
        at={(0.5,1.02)}, anchor=south,
        legend columns=-1,
        draw=none, fill=none,
        /tikz/every even column/.append style={column sep=0.30cm},
        font=\small,
    },
    legend image code/.code={
        \draw[#1, draw=black!55, line width=0.3pt]
            (0cm,-0.08cm) rectangle (0.28cm,0.14cm);
    },
]
\addplot+[draw=black!55, line width=0.3pt,
          fill=blue!55!cyan!80, fill opacity=0.85]
  coordinates {(USB-A,0.240) (USB-C,0.160) (RJ45,0.160) (HAN10E,0.127)
               (M12,0.153) (NEMA 1-15P,0.153) (RCA,0.153)
               (16mm Bar,0.220) (16mm Cyl,0.193)};
\addplot+[draw=black!55, line width=0.3pt,
          fill=orange!75, fill opacity=0.85]
  coordinates {(USB-A,0.127) (USB-C,0.193) (RJ45,0.220) (HAN10E,0.127)
               (M12,0.127) (NEMA 1-15P,0.067) (RCA,0.127)
               (16mm Bar,0.160) (16mm Cyl,0.067)};
\addplot+[draw=black!55, line width=0.3pt,
          fill=teal!60, fill opacity=0.85]
  coordinates {(USB-A,0.067) (USB-C,0.067) (RJ45,0.113) (HAN10E,0.000)
               (M12,0.000) (NEMA 1-15P,0.067) (RCA,0.067)
               (16mm Bar,0.067) (16mm Cyl,0.000)};
\addplot+[draw=black!55, line width=0.3pt,
          fill=red!65!magenta, fill opacity=0.85]
  coordinates {(USB-A,0.273) (USB-C,0.260) (RJ45,0.293) (HAN10E,0.360)
               (M12,0.420) (NEMA 1-15P,0.300) (RCA,0.353)
               (16mm Bar,0.127) (16mm Cyl,0.280)};
\addplot+[draw=black!55, line width=0.3pt,
          fill=violet!55, fill opacity=0.85]
  coordinates {(USB-A,0.180) (USB-C,0.220) (RJ45,0.160) (HAN10E,0.220)
               (M12,0.160) (NEMA 1-15P,0.260) (RCA,0.153)
               (16mm Bar,0.193) (16mm Cyl,0.127)};
\addplot+[draw=black!55, line width=0.3pt,
          fill=ForestGreen!75, fill opacity=0.85]
  coordinates {(USB-A,0.120) (USB-C,0.093) (RJ45,0.060) (HAN10E,0.120)
               (M12,0.133) (NEMA 1-15P,0.180) (RCA,0.147)
               (16mm Bar,0.247) (16mm Cyl,0.313)};
\legend{Align(pick), Engage (pick), Transport, Align (place), Engage (place), Success}
\end{axis}
\end{tikzpicture}
\caption{Assembly tasks (install).}
\label{fig:per-phase-failures-assembly}
\end{subfigure}

\vspace{6pt}

\begin{subfigure}{\textwidth}
\centering
\begin{tikzpicture}
\begin{axis}[
    width=\textwidth,
    height=4.8cm,
    ybar=0pt,
    bar width=3.4pt,
    ymin=0, ymax=0.75,
    ytick={0,0.2,0.4,0.6},
    ylabel={Fraction of rollouts},
    ylabel style={font=\small},
    symbolic x coords={USB-A, USB-C, RJ45, HAN10E, M12, NEMA 1-15P,
                       RCA, 16mm Bar, 16mm Cyl},
    xtick=data,
    x tick label style={
        font=\small,
        rotate=35,
        anchor=north east,
        yshift=-1pt,
        xshift=2pt,
    },
    enlarge x limits=0.07,
    tick align=outside,
    tick pos=left,
    ymajorgrids,
    major grid style={dashed, gray!35, line width=0.4pt},
    axis line style={gray!60},
]
\addplot+[draw=black!55, line width=0.3pt,
          fill=blue!55!cyan!80, fill opacity=0.85]
  coordinates {(USB-A,0.127) (USB-C,0.167) (RJ45,0.020) (HAN10E,0.067)
               (M12,0.207) (NEMA 1-15P,0.127) (RCA,0.213)
               (16mm Bar,0.067) (16mm Cyl,0.087)};
\addplot+[draw=black!55, line width=0.3pt,
          fill=orange!75, fill opacity=0.85]
  coordinates {(USB-A,0.067) (USB-C,0.047) (RJ45,0.153) (HAN10E,0.067)
               (M12,0.253) (NEMA 1-15P,0.067) (RCA,0.020)
               (16mm Bar,0.107) (16mm Cyl,0.140)};
\addplot+[draw=black!55, line width=0.3pt,
          fill=teal!60, fill opacity=0.85]
  coordinates {(USB-A,0.067) (USB-C,0.020) (RJ45,0.193) (HAN10E,0.000)
               (M12,0.067) (NEMA 1-15P,0.000) (RCA,0.067)
               (16mm Bar,0.093) (16mm Cyl,0.080)};
\addplot+[draw=black!55, line width=0.3pt,
          fill=red!65!magenta, fill opacity=0.85]
  coordinates {(USB-A,0.153) (USB-C,0.127) (RJ45,0.067) (HAN10E,0.193)
               (M12,0.087) (NEMA 1-15P,0.147) (RCA,0.113)
               (16mm Bar,0.067) (16mm Cyl,0.000)};
\addplot+[draw=black!55, line width=0.3pt,
          fill=violet!55, fill opacity=0.85]
  coordinates {(USB-A,0.067) (USB-C,0.000) (RJ45,0.047) (HAN10E,0.220)
               (M12,0.020) (NEMA 1-15P,0.020) (RCA,0.000)
               (16mm Bar,0.040) (16mm Cyl,0.000)};
\addplot+[draw=black!55, line width=0.3pt,
          fill=ForestGreen!75, fill opacity=0.85]
  coordinates {(USB-A,0.520) (USB-C,0.640) (RJ45,0.520) (HAN10E,0.460)
               (M12,0.360) (NEMA 1-15P,0.640) (RCA,0.600)
               (16mm Bar,0.647) (16mm Cyl,0.687)};
\end{axis}
\end{tikzpicture}
\caption{Disassembly tasks (remove).}
\label{fig:per-phase-failures-disassembly}
\end{subfigure}

\caption{Distribution of last-completed phase across precision assembly and disassembly tasks. Each group represents one connector; each bar shows the fraction of policy rollouts whose last successfully completed phase was the indicated phase, with the rightmost bar (\textcolor{ForestGreen!75!black}{green}) showing the fraction that completed all five phases (full task success).
Aggregated across all three policy families ($N{=}15$ rollouts per task per policy). .}
\label{fig:per-phase-failures}
\end{figure}

To characterize the empirical failure distribution, we ran 20 rollouts per task per policy across all three policy families, and for each rollout an annotator manually identified the phase at which the rollout first deviated from the expert trajectory and the corresponding failure mode. Aggregating across tasks and policies yields the per-phase failure frequencies reported in Figure~\ref{fig:per-phase-failures}, which we use to calibrate the recovery-episode collection: tasks and phases that fail more frequently in autonomous rollouts receive proportionally more recovery demonstrations.

Figure~\ref{fig:per-phase-failures} shows the distribution of the last successful phase across the three policy families we evaluate. Failures concentrate in the contact-rich Engage and place phases across all three architectures, with tighter-tolerance connectors exhibiting steeper progress drop-offs in late phases. This concentration motivates the recovery-data emphasis on contact-rich phases.

\paragraph{Recovery episodes.}
The recovery half of the dataset is grounded in the policies' own failure behaviour. We run each policy family autonomously on each task and identify the failure states, phases, and modes that occur in real rollouts. These define the target conditions for recovery collection: the operator reproduces each identified failure at the same phase and teleoperates the correction to task completion. The failure distribution is thus determined by trained policies, while the corrective behaviour is human-demonstrated.

\section{Analysis}
We analyze failure distributions across all three policy families (ACT, Diffusion Policy, $\pi_0$-FAST) on the 18-task suite, recording for each rollout the last successfully completed phase ($N{=}15$ rollouts per task per policy, 810  total). Aggregating across architectures yields four observations that inform both how recovery data should be calibrated and where phase-aware policy designs are most likely to help.

\paragraph{Failures concentrate in the object placement installation phases.}
Across all 9 installation tasks, Align (place) and Engage (place) together account for 54\% of all rollout failures, despite covering only two of the five phases. The Transport phase, by contrast, accounts for only 8\%. This asymmetry holds across all three policy families and reflects the mechanical structure of the tasks: the pick phases involve grasping a pre-positioned object in free space, whereas the place phases require sub-millimetre alignment to a mating fixture under contact. Recovery data is calibrated to the per-phase failure distribution therefore concentrates on contact-rich placement of assembly tasks.

\paragraph{Disassembly is substantially easier than assembly.}
Mean rollout success across the 9 install variants is 16\%, compared to 56\% across the 9 remove variants (Figure~\ref{fig:per-phase-failures}). The asymmetry is mechanical rather than learning-related: extraction is largely a free-space pull from a known grasp pose, while insertion couples grasping with sub-millimetre alignment to a tolerance fixture. This finding motivates the install/remove pairing in our task design. It lets the dataset isolate the contribution of insertion-direction precision while controlling for object identity and fixture geometry.

\paragraph{Geometric symmetry interacts with failure mode.}
Analyzing tasks by rotational symmetry reveals a clear trend in Engage-phase failure rates, where fine motor action mates the object with its receptacle. Objects with continuous symmetry (M12, RCA, 16~mm cylinder) achieve the lowest Engage failure rate at 14\% on average, followed by $C_1$ objects (USB-A, RJ45, HAN10E) at 19\%, and $C_2$ objects (USB-C, NEMA 1-15P, 16~mm bar) at 22\%. This trend reflects the increasing placement difficulty of keyed objects, which must be rotationally aligned to a single valid yaw before insertion.

\pgfmathsetseed{7}        
 
\definecolor{cACT}{HTML}{3B6FC4}
\definecolor{cDP} {HTML}{E07B2C}
\definecolor{cPI} {HTML}{2E7D46}

\pgfplotstableread[col sep=space]{
A1 A2 A3 A4 A5 A6  D1 D2 D3 D4 D5 D6  P1 P2 P3 P4 P5 P6
4  2  1  2  3  4   3  2  1  2  3  2   2  3  1  2  2  5
3  1  1  5  4  2   2  1  1  5  5  1   2  1  1  4  3  5
3  3  1  3  4  2   2  3  1  5  3  0   2  2  1  4  3  2
4  2  1  5  2  2   4  2  1  4  2  1   3  2  1  3  4  2
2  2  0  6  4  2   2  2  0  6  3  1   2  2  0  5  2  2
3  4  2  4  2  1   2  3  2  5  2  1   2  2  1  5  3  1
3  2  0  6  2  2   2  2  0  7  2  1   2  2  0  6  3  3
3  1  0  3  2  6   3  1  0  5  2  3   2  1  0  5  2  5
1  2  1  6  3  3   3  2  1  6  2  0   2  2  1  4  2  4
}\install
 
\pgfplotstableread[col sep=space]{
A1 A2 A3 A4 A5 A6  D1 D2 D3 D4 D5 D6  P1 P2 P3 P4 P5 P6
0  2  1  1  1  10  3  2  2  1  0  7   0  1  1  1  1  12
1  1  0  3  0  10  4  1  0  2  1  7   1  1  0  2  0  12
2  0  0  3  0  10  4  2  1  1  0  7   2  0  0  2  0  12
2  1  1  3  0  8   2  1  1  2  3  5   2  1  1  2  0  10
1  1  0  3  4  6   1  1  0  3  3  6   1  1  0  2  2  8
0  3  3  1  0  8   1  2  3  1  2  5   0  2  2  1  0  10
3  4  1  2  0  5   3  5  1  1  1  4   3  3  1  1  0  7
2  2  1  0  0  10  1  3  2  0  0  9   1  2  1  0  0  12
3  0  1  2  0  9   5  1  1  2  0  6   2  0  1  1  0  12
}\removetbl

\newcommand{\phbox}[4]{%
  \addplot[
    boxplot={draw position=#2, box extend=0.20, draw direction=y},
    color=#3, fill=#3, fill opacity=0.15,
    solid, line width=0.6pt, mark=none,
  ] table[y=#1] {#4};
}

\newcommand{\phpts}[5]{%
  \addplot[
    only marks, mark=#4, mark size=1.3pt,
    color=#3, fill=#3, mark options={draw=white, line width=0.25pt},
  ] table[x expr={#2+(rnd-0.5)*0.11}, y=#1] {#5};
}

\newcommand{\phgroup}[3]{%
  \phbox{A#1}{#2-0.22}{cACT}{#3}\phpts{A#1}{#2-0.22}{cACT}{*}{#3}%
  \phbox{D#1}{#2}{cDP}{#3}      \phpts{D#1}{#2}{cDP}{square*}{#3}%
  \phbox{P#1}{#2+0.22}{cPI}{#3} \phpts{P#1}{#2+0.22}{cPI}{triangle*}{#3}%
}
 
\pgfplotsset{
  phaseaxis/.style={
    boxplot/draw direction=y,
    xmin=0.45, xmax=6.7,
    xtick={1,2,3,4,5,6},
    xticklabels={Al(p),En(p),Tr.,Al(pl),En(pl),Success},
    xlabel={Last completed phase},
    ymajorgrids, grid style={densely dotted, gray!45},
    axis x line*=bottom, axis y line*=left,
    tick align=outside, tick style={semithick, black!55},
  },
}
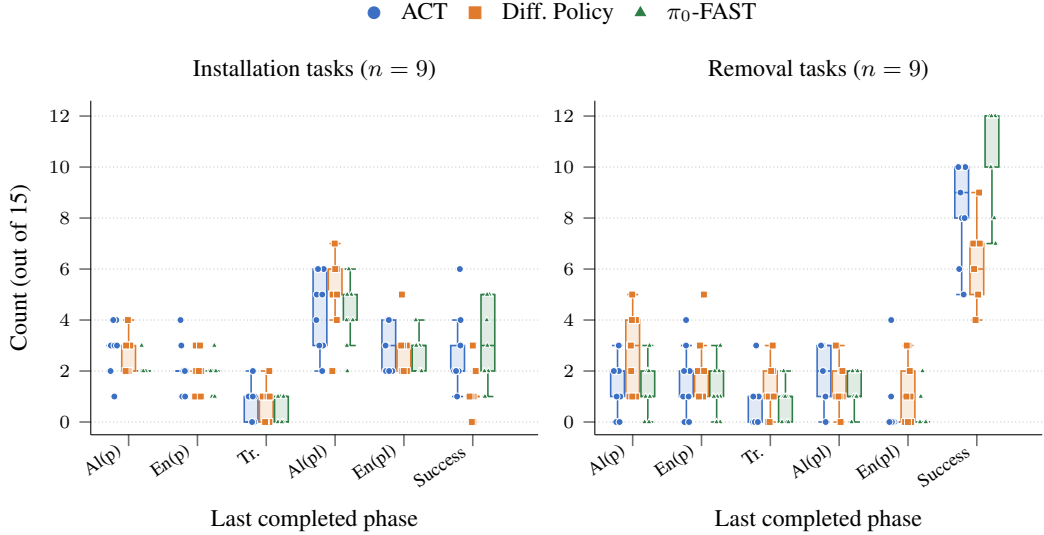
\begin{figure}[t]
\centering
\begin{tikzpicture}
\begin{groupplot}[
  group style={group size=2 by 1, horizontal sep=0.75cm},
  phaseaxis,
  width=7.5cm, height=6cm,
  xmax=6.95,
  ymin=-0.5, ymax=12.6, ytick={0,2,4,6,8,10,12},
  xticklabel style={font=\scriptsize, rotate=35, anchor=east, yshift=1pt},
  yticklabel style={font=\scriptsize},
  xlabel style={font=\small}, ylabel style={font=\small},
  title style={font=\small, yshift=-2pt},
]
\nextgroupplot[ylabel={Count (out of 15)}, title={Installation tasks ($n=9$)},
  legend style={draw=none, fill=none, at={(1.06,1.20)}, anchor=south,
                legend columns=3, column sep=6pt, font=\small},
  legend image post style={mark size=2.2pt}]
\addlegendimage{only marks, mark=*, color=cACT, fill=cACT}\addlegendentry{ACT}
\addlegendimage{only marks, mark=square*, color=cDP, fill=cDP}\addlegendentry{Diff.\ Policy}
\addlegendimage{only marks, mark=triangle*, color=cPI, fill=cPI}\addlegendentry{$\pi_0$-FAST}
\phgroup{1}{1}{\install}
\phgroup{2}{2}{\install}
\phgroup{3}{3}{\install}
\phgroup{4}{4}{\install}
\phgroup{5}{5}{\install}
\phgroup{6}{6}{\install}
 
\nextgroupplot[title={Removal tasks ($n=9$)}]
\phgroup{1}{1}{\removetbl}
\phgroup{2}{2}{\removetbl}
\phgroup{3}{3}{\removetbl}
\phgroup{4}{4}{\removetbl}
\phgroup{5}{5}{\removetbl}
\phgroup{6}{6}{\removetbl}
\end{groupplot}
\end{tikzpicture}
\caption{Phase-failure profiles for installation and removal tasks. Boxes show median and interquartile range with whiskers at $1.5\times$IQR; markers show all nine per-task counts. The install/remove asymmetry is visible at the Success endpoint, where removal counts are several times higher.}
\label{fig:phase-failure-both}
\end{figure}

\paragraph{Architecture-specific failure points are invisible under binary evaluation.}
Per-policy breakdowns (Table~\ref{fig:phase-failure-both}) reveal that ACT and Diffusion Policy achieve similar overall failure rates but diverge at the phase level. ACT struggles more in the Align phase on threaded fasteners (M12 install, 6/15 vs.\ 7/15), while Diffusion Policy stalls more often in the Engage phase on tight connectors (RJ45 install, 5/15 vs.\ 4/15). $\pi_0$-FAST achieves the highest success rate across both assembly and disassembly tasks. Notably, all three policies tend to fail at the same phases across tasks, underscoring the importance of targeted Align and Engage recovery demonstrations in the dataset.

\subsection{Future work and limitations}
\paragraph{Recovery policy design}
Our work aims to drive research in recovery policy design. Our failure recovery datasets enable this and encourage exploration of how recovery data can be used. The released annotations are intended to support several follow-up directions including training only on the corrective segment from $t_r$ onward, removing the failure segments $[t_f,t_r)$ while retaining the nominal approach, and integration of expert and recovery demonstrations. Determining which approach is most effective, is an important direction for future work.

\paragraph{Lab-controlled collection.}
All demonstrations were collected under consistent lighting, fixed camera placements, and a curated object set. This controlled setup reflects the scope of our benchmark, which focuses on precision manipulation and failure recovery rather than robustness to variations in lighting, background, or viewpoint. Our dataset provides limited visual diversity, and transfer to the clutter, occlusion, and workpiece variance of real-world assembly setups remains an open question. Extending data collection to more visually diverse environments and real manufacturing sites is a natural next step for our work.

\paragraph{Task coverage.}
Our task suite spans connector fastening and peg insertion across four precision tiers but excludes manipulation of flexible objects such as wire harnessing, PCB placement, and gasket seating. We view REBOOT as an extensible foundation: the phase decomposition and symmetry annotations generalize to any assembly task, and we welcome community contributions under the same schema.

\paragraph{Sensing modalities.}
Our work captures RGB, depth, and proprioception. For operations where vision is occluded by the gripper or the workpiece, force/torque and audio often carry the strongest engagement signal, and we do not claim these modalities are redundant for contact-rich assembly. A good direction for future work is include additional modalities as this will broaden the scope of learning approaches that can be applied.

\paragraph{Bi-manual manipulation}
Bi-manual demonstrations constitute only 83 of 2,160 demonstrations (3.8\%) in our dataset and are concentrated primarily in the \texttt{recovery\_install} tasks, where the non-dominant arm is used to reposition or reorient objects during recovery. The dataset therefore provides
limited coverage of sustained or coordinated bi-manual manipulation. Although the tasks can in principle be configured for execution with either arm, object initialization was standardized for right-arm execution to match the dominant hand of the teleoperators and to maintain consistency in tasks requiring sub-millimetre precision. This introduces a right-arm bias and limits evaluation of cross-arm generalization and left-arm execution.

\paragraph{Platform characteristics.}
Episodes were recorded within a fixed durations for protocol consistency across operators and tasks, so some contain trailing static segments after task completion. We distribute episodes untrimmed so that users may apply their own convention, and release a trimming utility that removes these segments.

\section{Conclusion}
The per-phase failure distribution reveals a structure that binary success obscures. Failures concentrate in placement, asymmetrically between insertion and extraction, and shift across phases as a function of mechanical symmetry. This structure is not a by-product of REBOOT but its organizing principle. For each task, recovery episodes are allocated across phases in proportion to the observed failure distribution during expert-only policy rollouts rather than uniformly, so that supervision is densest where policies actually fail.

RaC \citep{hu2025racrobotlearninglonghorizon} and RACER \citep{dai2024racerrichlanguageguidedfailure} establish that policies trained solely on successful demonstrations recover poorly when execution deviates from the expert distribution, and address this using human-in-the-loop intervention and data augmentation respectively. REBOOT takes a complementary approach, recording failures and their corrections systematically so that they can be studied offline. Our per-phase analysis shows that failure modes often recur across tasks with similar mechanical structure, suggesting that recovery strategies can be transferred across related tasks. 

While multi-task and meta-learning have successfully exploited shared structure in expert demonstrations \citep{finn2017modelagnosticmetalearningfastadaptation, duan2017oneshotimitationlearning}, an equivalent representation of failure has been largely unavailable. REBOOT provides this: 2,160 teleoperated demonstrations with annotated phase boundaries, failure and recovery timestamps, failure-mode labels, and a bi-manual interaction flag, as well as the per-phase evaluation protocol used to produce the failure distributions we report. This gives both the data and the measurement framework against which failure-aware methods can be developed and compared.

\newpage
\begin{ack}
We thank the Natural Sciences and Engineering Research Council of Canada (NSERC), CIFAR, and Schulich Momentum for funding and supporting this research.
We also thank Digital Research Alliance of Canada and Research Computing Services at the University of Calgary for computational resources.
\end{ack}

\medskip
\newpage
{
\small
\bibliographystyle{unsrtnat}
\bibliography{ref}
}

\newpage
\appendix

\section{Appendix}

\subsection{Hardware and software setup}
\paragraph{Hardware:}
This dataset was collected using Trossen Robotics' WidowX AI bi-manual tele-operation system, at a 30 Hz control frequency by a single human operator. The system consists of two 6-DOF robot follower arms, each paired with a corresponding leader arm fitted with a hand grip in place of a gripper for teleoperation. The arms are mounted on an aluminum frame clamped to a stable table. Each arm has a maximum payload capacity of 1.5 kg. The sensors comprise four Intel RealSense D405 RGB-D cameras: one wrist-mounted camera on each gripper, a fixed overhead camera providing a birds-eye view of the workspace, and a fixed front-facing camera at table height. All cameras record synchronized RGB video and single-channel uncompressed depth feed at 640$\times$480 resolution and 30 fps, encoded using ffmpeg. 

Joint positions for both arms are recorded in radians for the six joints per robot arm and metres for displacement of the parallel gripper. Cartesian end-effector poses are not logged directly but are fully determined by the recorded joint positions, since both arms are rigidly mounted and are not moved between episodes. End-effector pose of each arm at every timestep can be derived from the recorded joint angles with the publicly available URDF for the platform; LeRobot provides forward-kinematics utilities for this conversion. 

\begin{figure}[H]
\centering
\begin{lstlisting}[language=Python, basicstyle=\footnotesize\ttfamily,
                   frame=single, breaklines=true]
from lerobot.model.kinematics import RobotKinematics
 
kinematics = RobotKinematics(
    urdf_path="./widowx_ai/widowx_ai.urdf",
    target_frame_name="ee_gripper_link",
    joint_names=["joint_0", "joint_1", "joint_2",
                 "joint_3", "joint_4", "joint_5"],
)
 
# joint positions (rad) -> end-effector pose in the arm base frame
ee_pose = kinematics.forward_kinematics(joint_positions)
\end{lstlisting}
\caption{Recovering Cartesian end-effector poses from the recorded joint
positions using the published kinematic model. A batch script that applies this
to every episode in the release is provided in our codebase.}
\label{fig:fk_snippet}
\end{figure}

\paragraph{Workspace setup:}
The taskboard is placed in the centre of a wood-finish work surface, directly in front of the robot. Overhead LED panels provide consistent lighting across all data collection sessions. All recordings take place in a single laboratory room with a blacked-out background, which eliminates incidental foot traffic from camera views and preserves participant privacy.

We used the recommended worktable from Trossen Robotics for the WidowXAi stationary setup; the ULINE Industrial Packing Table - 48 x 30", maple top with square edge. A rigid aluminum frame is clamped to the workspace and has mounting points for leader and follower robot arms, overhead and low-mounted cameras and a touch screen for control. The left and right follower arms are bolted to the left and right sections of the frame, with their workspaces separated by 84 cm across the width of the table, and are centered in their depth 37 cm from both near and far edges. This setup with the frame and the table creates a workspace of 84 x 74 cm. The leader robot arms are on the same frame as the follower robot arms on the workspace, extended 28 cm beyond the frame edge nearest to the teleoperator. Assembly instructions, technical drawings and kinematic models for the WidowX AI platform are publicly available from \href{https://drive.google.com/drive/folders/1bVi37QH0NhnrXrNQ6zcDeJ-FLGykucKv?usp=sharing}{Trossen Robotics' public folder}. 

\paragraph{Camera setup:}
We use 4 Intel Realsense D405 RGB-D cameras: one wrist mounted camera on each robot arm just above the parallel grippers, one overhead camera mounted on a frame giving a top-down view of the workspace at a height of 102.5 cm above the workspace, and one low mounted, forward facing camera at the base of the workspace on the frame edge nearest the teleoperator, centred laterally (42 cm from each side) closest to the teleoperator oriented toward the board to capture the insertion targets in profile. Technical drawings and assembly instructions for the frame and camera positioning are publicly available on the Trossen Robotics website.

\paragraph{Home configuration:}
At the beginning of every teleoperated demonstration, each robot arm moves from rest to a fixed home configuration: joint 0: 1.00 rad, joint 1:0.5 rad, joint 2: 0.5 rad, joint 3: 0 rad, joint 4: 0 rad, joint 5: 0 rad, parallel gripper: 0 m. These follow the manufacturer’s convention where they use radians for joint positions and meters for the parallel gripper displacement and can be converted to end-effector poses using the WidowXAI published kinematic model.

\paragraph{Taskboard and objects:}
Twelve of our 18 tasks use the NIST Assembly Task Board \#1, for which fabrication files, assembly instructions, and ordering information are publicly available from NIST(Kimble et al, “Benchmarking Protocols for Evaluating Small Parts Robotic Assembly Systems”), including the option to order a replica or 3D-print the board and objects. The remaining six tasks (RCA connector install/remove, USB-C install/remove, NEMA 1-15P install/remove) are our own extension of the board, built entirely from off-the-shelf components: standard connector cables cut approximately 2 in above the terminal, with the female end held in a tabletop mini-vise and the male end supported upright.

\paragraph{Board and object initialization:} For the twelve NIST task board tasks, the NIST taskboard \#1 is mounted at a fixed initial position on the workspace, suspended 2 cm off the surface by screw-on inserts, and offset from the left of the workspace by 28 cm and from the bottom of the workspace by 14 cm across for the twelve tasks that use the For each distinct task, the task board is reoriented in 90 degree steps such that the goal destination of objects are in a reasonable working range of the right robot arm, and also remains in the view of the front facing low mounted camera. Objects are placed in the center of a plastic container(16 cm by 22 cm) and initialized on the right of the taskboard. For the 6 tasks which use the mini-vise clamp, the clamp is positioned on the workspace at a 39 cm offset from the left side of the workspace, and 27 cm off from the bottom of the workspace. Objects are placed at the centre of a plastic container and initialized within the remaining 35 cm × 15 cm area to the left of the clamp, at positions within reach of the right arm.

\paragraph{Bill of materials.}
Table~\ref{tab:bom} lists the off-the-shelf components used to build the six custom tasks that extend the NIST board. All items were obtained from general retail suppliers and are listed with the specific products used, so that the setup can be replicated without fabrication. The mini-vise is secured to the work surface with double-sided mounting tape to keep its position fixed between episodes.
 
\begin{table}[H]
\centering
\small
\caption{Bill of materials for the six custom tasks extending the NIST
Assembly Task Board~\#1. Fabrication files and ordering information for the
task board itself are available from NIST~\citep{NIST2022}.}
\label{tab:bom}
\setlength{\tabcolsep}{5pt}
\renewcommand{\arraystretch}{1.15}
\begin{tabularx}{\textwidth}{@{} l X l @{}}
\toprule
\textbf{Component} & \textbf{Product used} & \textbf{Used for} \\
\midrule
Mini-vise &
  Tabletop drill-press vise with rubber soft jaws,
  \SIrange{0}{63}{\milli\meter} clamping range &
  Holding the female connector fixed \\
RCA connectors &
  2\,RCA-to-2\,RCA gold-plated audio extension cable
  (male and female ends), \SI{0.9}{\meter} &
  RCA install / remove \\
USB-C connectors &
  USB-C 3.2 Gen2 male-to-female extension cable,
  \SI{20}{Gbps} &
  USB-C install / remove \\
NEMA 1-15P (male) &
  Two-prong polarized AC power cord, 18~AWG, \SI{0.5}{\meter} &
  NEMA 1-15P install / remove \\
NEMA 1-15P (female) &
  Polarized two-prong extension cord, 16~AWG, \SI{3.0}{\meter} &
  NEMA 1-15P install / remove \\
Mounting tape &
  Double-sided indoor mounting tape &
  Fixing the mini-vise to the work surface \\
Object container &
  Rectangular food container,
  \SI{16}{\centi\meter} $\times$ \SI{22}{\centi\meter} &
  Object initialization region \\
\bottomrule
\end{tabularx}
\end{table}
 
Each connector cable is cut approximately 5 cm above the terminal to get a separate male and female piece. The female end is clamped in the mini-vise; the male end is positioned upright, supported by a small block of foam so that it presents a consistent grasp pose at the start of each episode.

\begin{figure}[H]
\centering
\includegraphics[width=0.5\textwidth]{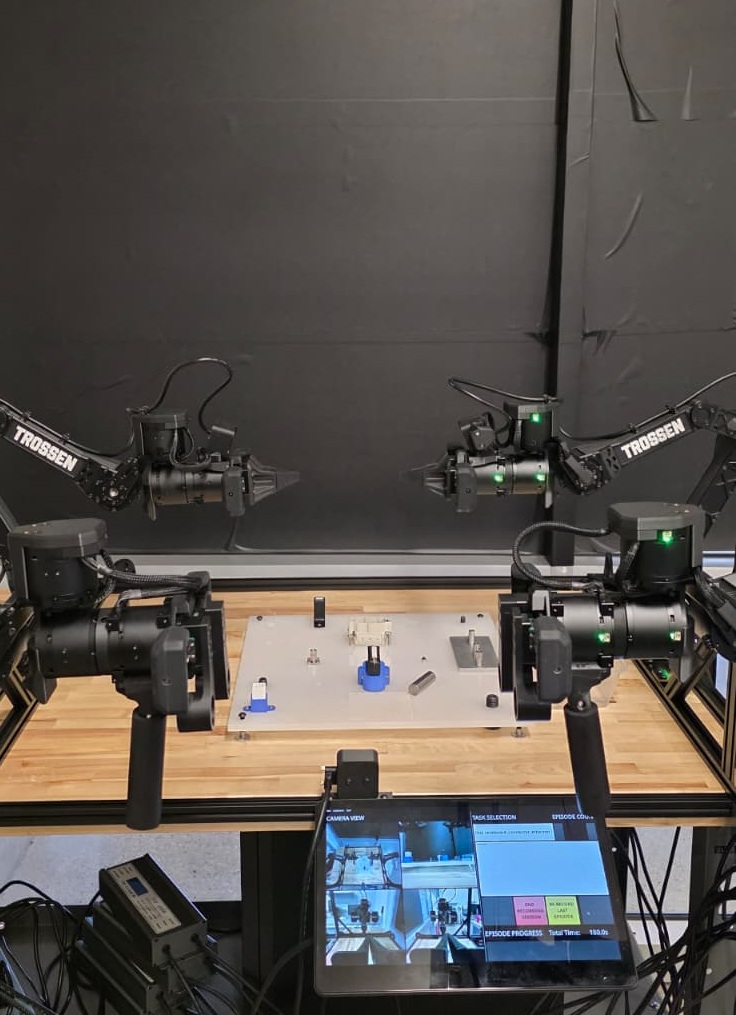}
\caption{Hardware and teleoperation setup for REBOOT data collection. Two Trossen WidowX AI 6-DOF follower arms (lower) manipulate connectors on a NIST Assembly Task Board~\#1(centre); two matching leader arms (upper) provide kinesthetic teleoperation. The control workstation (foreground) runs the LeRobot recording interface, displaying live feeds from all four
RGB-D cameras alongside task and episode controls.}
\label{fig:hardware_setup}
\end{figure}

\paragraph{Software setup}
Dataset recording, policy training, and rollout are executed through Trossen AI's fork of the open-source LeRobot framework(v0.4.3)~\cite{cadene2024lerobot}, available at \url{https://github.com/TrossenRobotics/lerobot_trossen}. This fork was further extended with PR~2604 from the upstream LeRobot repository to enable depth data recording and storage~(\url{https://github.com/huggingface/lerobot/pull/2604}). An additional script for editing episode length can be found in our codebase.(\url{https://github.com/anon-robo-account/REBOOT}) 

\subsection{Dataset structure and access}
\label{app:dataset_structure}

REBOOT is publicly hosted on HuggingFace at
\url{https://huggingface.co/REBOOT26/datasets}, distributed under the Creative Commons Attribution 4.0 International (CC-BY-4.0) license. Each expert and recovery task is hosted as an independent dataset on the REBOOT. All trajectories are stored in the LeRobot v3 dataset format, organized into three top-level folders, \texttt{data}, \texttt{meta}, and \texttt{video},as described below.

\paragraph{\texttt{Data}}
Parquet files(.parquet) containing joint positions and gripper states for both arms at 30~Hz, synchronized with video timestamps. Depth streams are stored in the same parquet files as raw
single-channel arrays, with four streams per episode aligned with the corresponding RGB frames.

\paragraph{\texttt{Meta}}
JSON files containing per-episode and per-task annotations:
\begin{itemize}\itemsep2pt
\item \textbf{Task metadata:} task name, install/remove variant, rotational symmetry group ($C_1$, $C_2$, or continuous), precision tier, mating clearance, and natural-language task description.
\item \textbf{Phase annotations:} timestamp boundaries
  $(\tau_0,\dots,\tau_5)$ for the five phases of each trajectory. For recovery episodes, additional fields record the originating failure phase $k^\star$ and the categorical failure mode $m \in \{$misalignment, slip, premature release, jamming, off-axis
  collision$\}$. The annotation schema is illustrated in Figure~\ref{fig:json_example}, which shows a recovery episode for the RCA install task. Each episode carries the five phase boundaries and, for recovery episodes, the originating failure phase and mode that link the episode back to the empirical failure distribution it was sampled from.
\end{itemize}

\begin{figure}[H]
\centering
\begin{lstlisting}[language=json]
{
  "repo_id": "REBOOT26/rca_recovery_install",
  "task_id": "rca-install",
  "episode_index": "09",
  "task_description": "Pick the RCA connector and install it on the hub",
  "episode_type": "recovery",
  "duration_frames": 897,
  "fps": 30,
  "phase_boundaries": {
    "tau_0": 0,
    "tau_1": 210,
    "tau_2": 450,
    "tau_3": 540,
    "tau_4": 570,
    "tau_5": 897
  },
  "phase_names": [
    "Align(pick)",
    "Engage(pick)",
    "Transport",
    "Align(place)",
    "Engage(place)"
  ],
  "failure": {
    "originating_phase": 2,
    "failure_mode": "slip",
    "t_f": 210,
    "t_r": 420,
    "t_s": 612,
    "language_description": "Failed to grasp RCA connector; operator realigned gripper and completed the task."
  },
  "operator_id": "op-01",
  "collection_session": "2026-04-15"
}
\end{lstlisting}
\caption{Example phase and failure annotation for a recovery episode. The originating failure (slip) occurred during Engage(pick) at frame 210 ($t_f$); the operator initiated recovery at frame 420 ($t_r$) and returned the task to a recoverable state at frame 612 ($t_s$), completing it successfully.}
\label{fig:json_example}
\end{figure}

\paragraph{\texttt{Video}}
Four MP4(.mp4) RGB streams per episode at 640$\times$480, 30~fps, stored in separate sub-folders:
\begin{itemize}\itemsep2pt
\item \texttt{observation.images.cam\_high} --- overhead view
\item \texttt{observation.images.cam\_low} --- front-facing view
\item \texttt{observation.images.cam\_left\_wrist} --- left arm
  wrist view
\item \texttt{observation.images.cam\_right\_wrist} --- right arm
  wrist view
\end{itemize}

The dataset can be visualized using the LeRobot rerun-based SDK visualizer or the publicly accessible online tool at \url{https://huggingface.co/spaces/lerobot/visualize_dataset}.
Episode data follows the LeRobot v3 schema for direct compatibility with downstream training pipelines.
\begin{figure}[h]
    \centering
    \includegraphics[width=\linewidth]{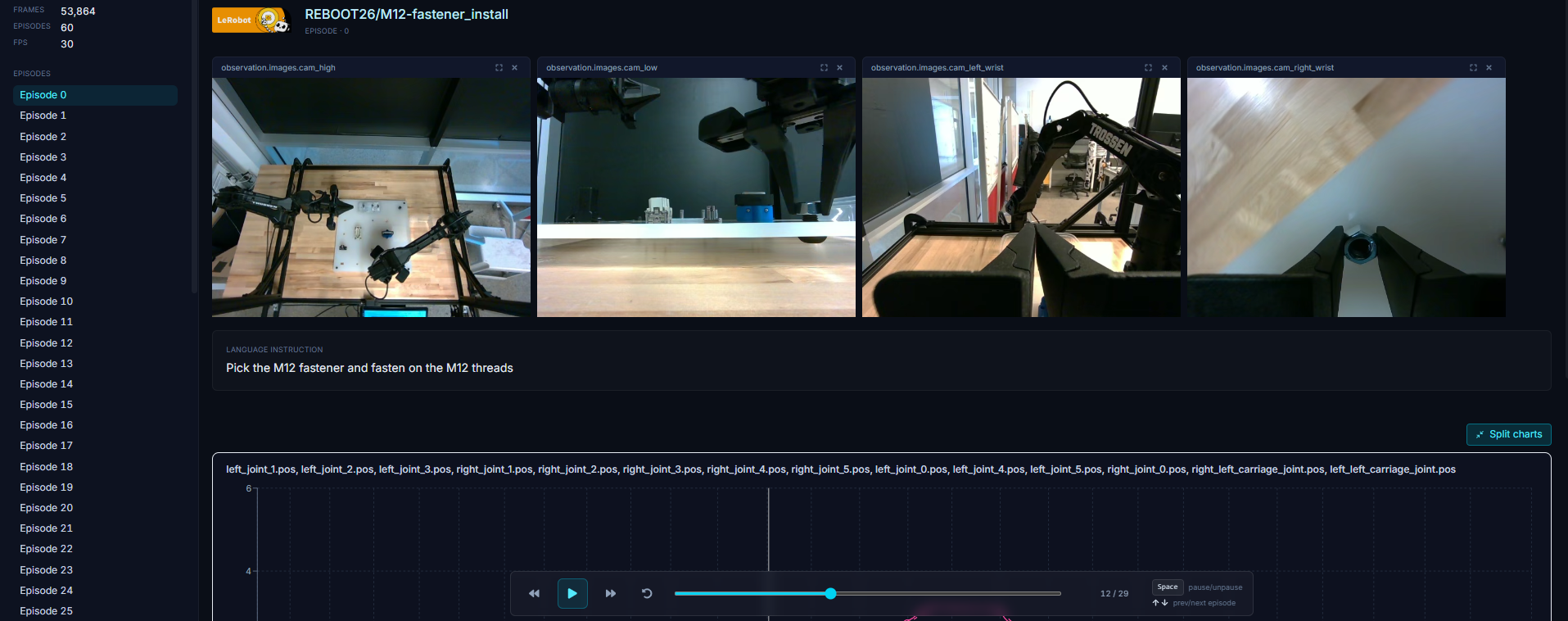}
    \caption{Screenshot of the LeRobot online dataset visualizer showing the \texttt{M12\_fastener\_install} task. Four synchronized RGB camera streams are displayed: overhead (\texttt{cam\_high}), front-facing (\texttt{cam\_low}), left-wrist (\texttt{cam\_left\_wrist}), and right-wrist (\texttt{cam\_right\_wrist}). The language instruction, joint position time series, and episode list are also shown.}
    \label{fig:dataset_visualizer}
\end{figure}

\begin{figure}[H]
    \centering
    \includegraphics[width=\linewidth]{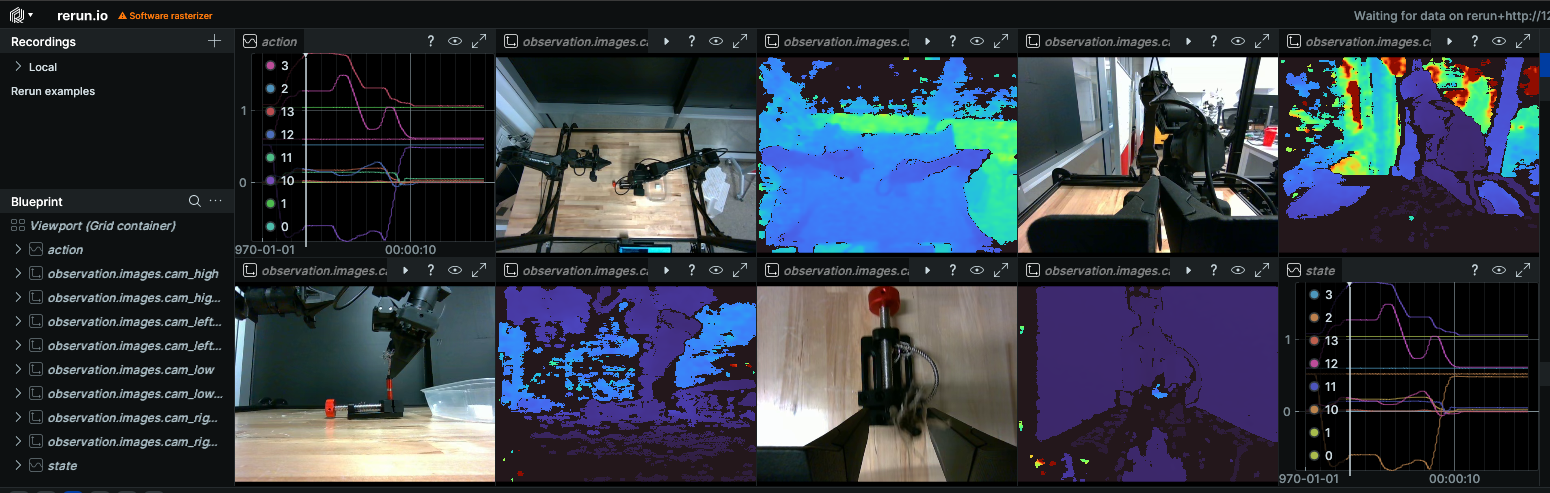}
    \caption{Screenshot of the Rerun SDK visualizer showing a recorded REBOOT episode. The interface displays synchronized RGB and depth streams from all four cameras (\texttt{cam\_high}, \texttt{cam\_low}, \texttt{cam\_left\_wrist}, \texttt{cam\_right\_wrist}), with depth rendered as false-colour maps alongside their corresponding RGB views. Action and state time series are shown in the left and right panels respectively, with joint indices colour-coded across the timeline.}
    \label{fig:rerun_visualizer}
\end{figure}

\subsection{Annotation protocol}
\label{app:annotation}

Phase boundaries and failure modes were annotated by a single graduate student (the first author) who collected the dataset and reviewed every episode post-hoc. We document the protocol below to support reproducibility and to clarify the operational definitions behind each annotation.

\paragraph{Phase boundary annotation.}
Each episode is partitioned into the five phases by four interior boundary frames $\tau_1, \tau_2, \tau_3, \tau_4$, with $\tau_0 = 1$ and $\tau_5 = T$ fixed at the episode endpoints. Boundaries were identified by replaying each episode at reduced playback speed and labelling the first frame at which the kinematic event defining the next phase had clearly begun. The defining events are:

\begin{itemize}\itemsep2pt
\item $\tau_1$ \textbf{(Align $\to$ Engage, pick):} first frame at which the gripper has reached its target grasp pose and begins closing on the source object. \item $\tau_2$ \textbf{(Engage $\to$ Transport, pick):} first frame at which the grasped object loses contact with its source fixture (lift-off).
\item $\tau_3$ \textbf{(Transport $\to$ Align, place):} first frame at which the held object enters the placement workspace and the gripper begins fine-positioning toward the goal pose.
\item $\tau_4$ \textbf{(Align $\to$ Engage, place):} first frame at which the held object makes contact with the goal fixture or begins seating motion.
\end{itemize}

Boundaries are annotated at single-frame resolution and stored as integer frame indices in each episode's JSON annotation file (Figure~\ref{fig:json_example}).

\paragraph{Failure and recovery definitions}
The failure and recovery events recorded for each recovery episode are defined with respect to the phase sequence above. For a task with ordered phases $p_1,\dots,p_K$, let $S_k$ denote the binary success predicate for phase $p_k$.

\begin{itemize}\itemsep2pt
\item \textbf{Failure onset $t_f$:} the first timestep at which $S_k$ switches to zero, that is, the first timestep from which task progression requires a corrective action outside the expert action distribution for $p_k$.
\item \textbf{Recovery initiation $t_r$:} the first timestep of the corrective action taken in response to the failure at $t_f$.
\item \textbf{Recovery success $t_s$:} the first timestep at which $S_k$ returns to one and execution proceeds to $p_{k+1}$ without further corrective intervention.
\end{itemize}
 
We derive the recovery lag for each episode as $t_r - t_f$. Although $S_k$ and the corresponding onset conditions are defined at the level of phase semantics, we additionally specify a criteria for each failure mode to ensure that annotations are reproducible from observable states rather than subjective annotator judgement. For example, a misalignment failure is annotated at the first frame in which the object is displaced by more than 5 mm from the insertion axis, while recovery is declared once the object returns to within 1 mm of the axis. 
 
Bi-manual recovery actions are annotated within the phase in which the failure occurred, with phase progression resuming once the dominant arm re-engages the object. Annotating them as separate phases would fragment the phase sequence and prevent direct comparison between expert and recovery episodes.

\paragraph{Failure mode annotation.}
For recovery episodes, the originating failure is annotated with two attributes: the phase $k^\star \in \{1,\dots,5\}$ during which the failure occurred, and the categorical mode
$m \in \{$misalignment, slip, premature release, jamming, off-axis collision$\}$. The operational definitions used during annotation are:

\begin{itemize}\itemsep2pt
\item \textbf{Misalignment:} end-effector pose deviates from the target alignment axis beyond the connector's clearance tolerance, preventing engagement. Annotated when the gripper closes on a pose visibly offset from the object centroid, or when the held object fails to enter its mating port due to translational or rotational offset.
\item \textbf{Slip:} the held object moves within or escapes the gripper after grasp, typically during transport or insertion. Annotated when post-grasp object pose changes are visible relative to the gripper fingers.
\item \textbf{Premature release:} the gripper opens before the object is fully seated or transferred, leaving the assembly incomplete. Annotated when the gripper command transitions from closed to open while the object is still in transit or only partially seated.
\item \textbf{Jamming:} the held object binds against its mating port mid-insertion, with insertion progress halted. Annotated when the held object enters the goal fixture but stops short of full seating with no further forward motion.
\item \textbf{Off-axis collision:} the arm or held object contacts surrounding fixtures during transport or retraction, displacing the workpiece or triggering a safety stop. Annotated when contact with fixtures other than the source or goal is visible.
\end{itemize}

In cases where multiple modes co-occurred (e.g., misalignment followed by jamming), annotators recorded the mode that initiated the failure cascade. Annotations were performed using a custom Python interface used in parallel with the LeRobot dataset visualizer for direct logging of phase boundary timestamps and failure mode labels into the per-episode JSON files described in Section~\ref{app:dataset_structure}.

\subsection{Per-task data statistics}
\label{app:per_task_stats}
Table~\ref{tab:per_task_counts} reports per-task statistics including expert and recovery episode counts, mean episode duration, and the phase distribution of induced failures within each task's recovery set. The recovery distribution is calibrated to the empirical failure distribution measured from autonomous policy rollouts on the corresponding task (Section~\ref{app:failure_calibration}).

\begin{table}[H]
\centering
\caption{Per-task episode counts and recovery-episode failure distributions. Phase columns sum to the recovery total per row. (Sym -- Rotational symmetry, Tier -- Clearances, Exp -- Expert demonstrations, Dur -- Duration, Al(p) -- Align(Place), En(p) -- Engage(Place), Tr. --Trasport, Al(pl) -- Align(place), En(pl) -- Engage(Place))  }
\label{tab:per_task_counts}
\setlength{\tabcolsep}{4pt}
\begin{tabular}{lcccccccccc}
\toprule
\textbf{Task} & \textbf{Sym.} & \textbf{Tier} & \textbf{Exp.} &
\textbf{Rec.} & \textbf{Dur. (s)} &
\textbf{Al(p)} & \textbf{En(p)} & \textbf{Tr.} & \textbf{Al(pl)} & \textbf{En(pl)} \\
\midrule
USB-A install      & C1 & Mod.  & 60 & 60 & 30 & 16 & 9 & 4 & 19 & 12 \\
USB-A remove       & C1 & Mod.  & 60 & 60 & 15 & 16 & 8 & 8 & 20 & 8 \\
USB-C install      & C2 & Tight & 60 & 60 & 30 & 11 & 12 & 4 & 18 & 15 \\
USB-C remove       & C2 & Tight & 60 & 60 & 15 & 28 & 8 & 4 & 20 & 0 \\
RJ45 install   & C1 & Tight & 60 & 60 & 30 & 10 & 14 & 7 & 19 & 10 \\
RJ45 remove    & C1 & Tight & 60 & 60 & 15 & 3 & 20 & 24 & 8 & 5 \\
HAN10E install     & C1 & Loose & 60 & 60 & 30 & 9 & 9 & 0 & 26 & 16 \\
HAN10E remove      & C1 & Loose & 60 & 60 & 30 & 7 & 7 & 0 & 22 & 24 \\
M12 fastener install & C2 & Mod. & 60 & 60 & 30 & 11 & 9 & 0 & 30 & 10 \\
M12 fastener remove  & C2 & Mod. & 60 & 60 & 30 & 20 & 25 & 6 & 7 & 2 \\
NEMA 1-15P install & C1 & Mod.  & 60 & 60 & 30 & 10 & 5 & 5 & 21 & 19 \\
NEMA 1-15P remove  & C1 & Mod.  & 60 & 60 & 15 & 21 & 11 & 0 & 24 & 4 \\
RCA install        & Cont. & Int. & 60 & 60 & 30 & 11 & 9 & 5 & 25 & 10 \\
RCA remove         & Cont. & Int. & 60 & 60 & 15 & 31 & 3 & 9 & 17 & 0 \\
16mm Bar install   & C2 & Mod.  & 60 & 60 & 30 & 17 & 12 & 5 & 10 & 16 \\
16mm Bar remove    & C2 & Mod.  & 60 & 60 & 15 & 11 & 18 & 16 & 10 & 5 \\
16mm Cyl install   & Cont. & Mod. & 60 & 60 & 30 & 17 & 6 & 0 & 25 & 12 \\
16mm Cyl remove    & Cont. & Mod. & 60 & 60 & 15 & 17 & 28 & 15 & 0 & 0 \\
\midrule
\textbf{Total}     & --  & --   & \textbf{1{,}080} & \textbf{1{,}080} & -- & -- & -- & -- & -- & -- \\
\bottomrule
\end{tabular}
\end{table}

\subsection{Failure calibration protocol}
\label{app:failure_calibration}
Recovery episodes are calibrated to match the empirical failure distribution observed during autonomous policy rollouts. The calibration protocol is as follows:

\begin{enumerate}\itemsep2pt
\item Train a baseline ACT, Diffusion policies and $\pi_0$-FAST on the expert split for each task ($N=60$ demonstrations, 24 GPU-hours per task).
\item Evaluate the trained policies autonomously for $R=15$ rollouts per task with randomized initial object pose within a 3-6cm radius of the nominal start. 
\item For each rollout, an annotator manually identifies the phase $k^\star$ at which the rollout first deviates from the expert trajectory and the corresponding failure mode $m$.
\item Aggregate the empirical phase-failure distribution $\hat{p}(k^\star \mid text{task})$ across the $R$ rollouts.
\item Collect 60 recovery episodes per task with the per-phase distribution matched to $\hat{p}(k^\star \mid \text{task})$.
\end{enumerate}

For each recovery episode, the operator deliberately induces the sampled failure mode at the sampled phase, then teleoperates the robot from the failure state to task completion. Table~\ref{tab:phase_examples} shows examples of failed trajectories and recovery actions inititaed.

\begin{table}[H]
\centering
\small
\caption{Qualitative examples of successful and failed trajectories across the five phases of our task decomposition. Each phase exhibits a characteristic failure mode, and the language descriptions accompany
each frame to provide grounded supervision for vision-language recovery policies.}
\label{tab:phase_examples}
\renewcommand{\arraystretch}{0.5}
\setlength{\tabcolsep}{6pt}
\begin{tabularx}{\textwidth}{@{} l l X C{3cm} @{}}
\toprule
Phase & Outcome & Language description & Picture \\
\midrule

\multirow{2}{*}{Align(Pick)}
 & Success & End-effector moves from the home pose to  grasp range of the target connector
 & \includegraphics[width=2.5cm]{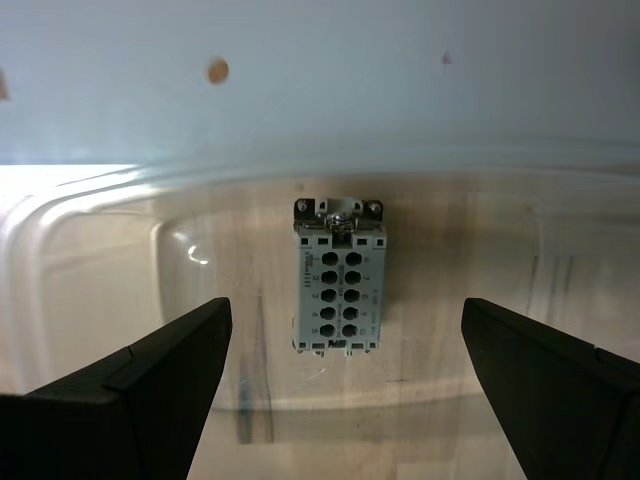} \\
\cmidrule(l){2-4}
 & Failure & End-effector overshoots the target or stalls in free space
 & \includegraphics[width=2.5cm]{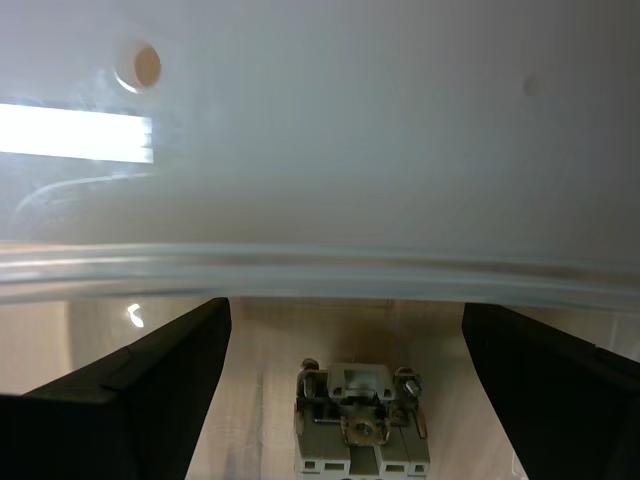} \\
\midrule

\multirow{2}{*}{Engage(Pick)}
 & Success & Gripper perfectly grasps object
 & \includegraphics[width=2.5cm]{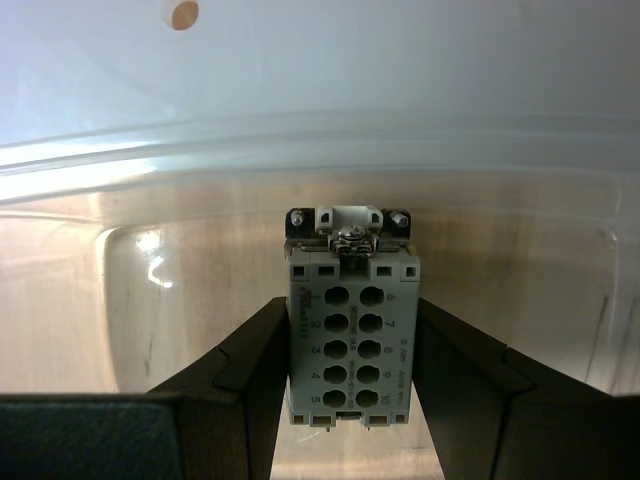} \\
\cmidrule(l){2-4}
 & Failure & Gripper closes on a misaligned pose missing object.
 & \includegraphics[width=2.5cm]{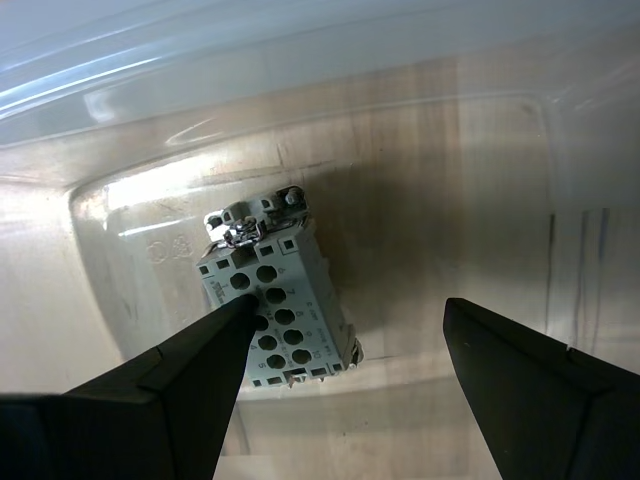} \\
\midrule

\multirow{2}{*}{Transport}
 & Success & Object remains grasped as the arm moves to the goal location
 & \includegraphics[width=2.5cm]{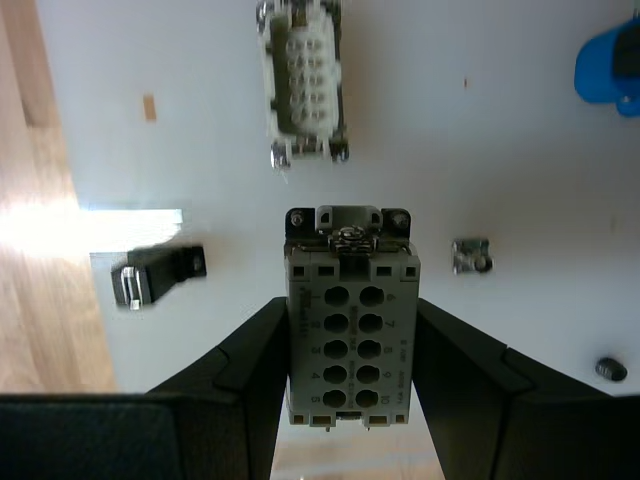} \\
\cmidrule(l){2-4}
 & Failure & Arm drifts off-axis or gripper for reduces and drops the object
 & \includegraphics[width=2.5cm]{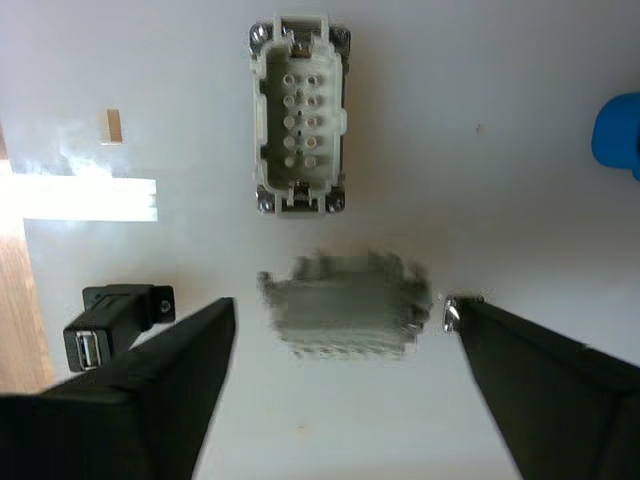} \\
\midrule

\multirow{2}{*}{Align(Place)}
 & Success & Connector axis misaligned mating port
 & \includegraphics[width=2.5cm]{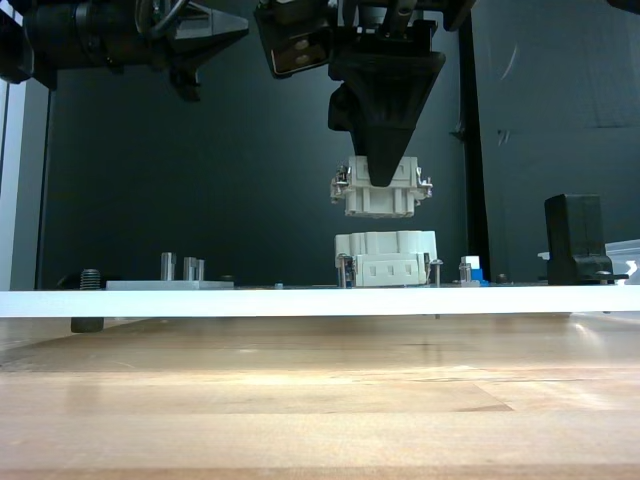} \\
\cmidrule(l){2-4}
 & Failure & Connector axis misaligned with center of mating port
 & \includegraphics[width=2.5cm]{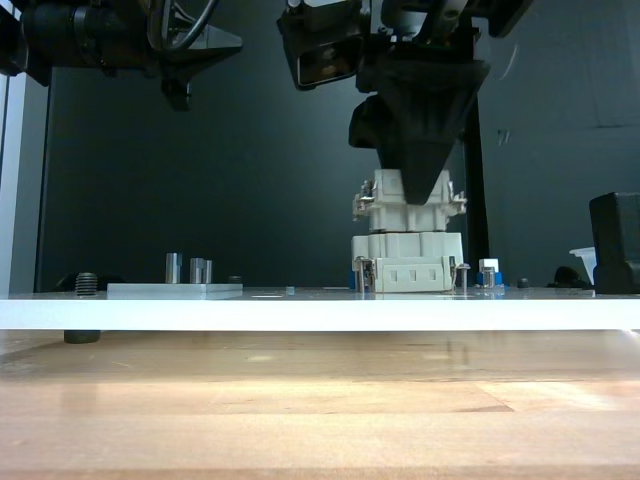} \\
\midrule

\multirow{2}{*}{Engage(Place)}
 & Success & Connector enters its mating port completely
 & \includegraphics[width=2.5cm]{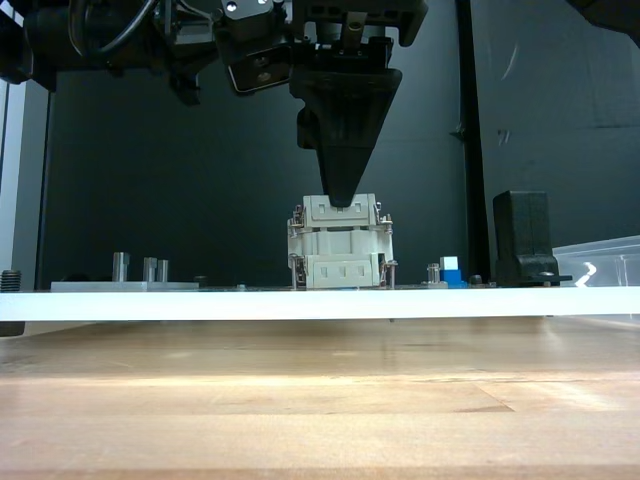} \\
\cmidrule(l){2-4}
 & Failure & Connector jams partway through insertion
 & \includegraphics[width=2.5cm]{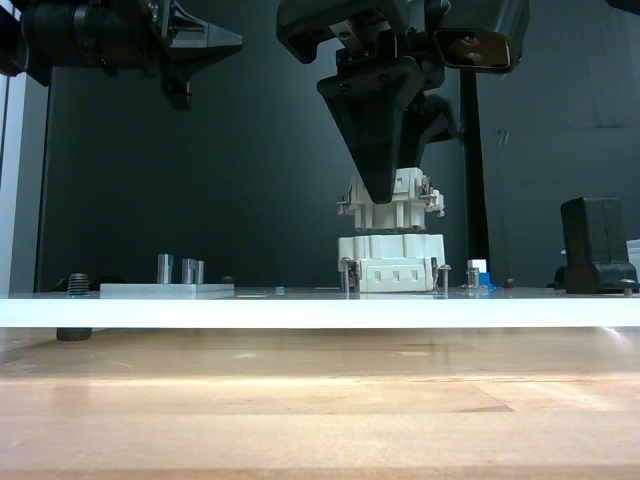} \\
\midrule
\bottomrule
\end{tabularx}
\end{table}

\subsection{Bi-manual episode inventory}
 
Bi-manual manipulation occurs when a failure displaces an object outside the dominant right arm's reachable workspace, requiring the non-dominant left arm to reposition, reorient, or retrieve it before the task can continue. It is confined to the recovery split, specifically to the \texttt{recovery\_install} tasks. In the \texttt{recovery\_remove} tasks, failures occur within the dominant arm's reach and reorientation is not required for disassembly. Expert demonstrations are single-armed throughout.
 
Table~\ref{tab:bimanual_counts} lists the episodes containing bi-manual interaction for each task. 
 
\begin{table}[H]
\centering
\small
\caption{Episodes containing bi-manual interaction, by task. All occur in the recovery split of installation tasks. Episode indices are as released.}
\label{tab:bimanual_counts}
\setlength{\tabcolsep}{5pt}
\renewcommand{\arraystretch}{1.2}
\begin{tabularx}{\textwidth}{@{} l c X @{}}
\toprule
\textbf{Task (recovery, install)} & \textbf{Count} & \textbf{Episode indices} \\
\midrule
RJ45          & 23 & 2, 3, 4, 10, 11, 17, 23, 25, 26, 27, 28, 29, 30, 31, 34,
                     41, 42, 48, 49, 50, 53, 55, 58 \\
16\,mm bar    & 15 & 1, 8, 9, 10, 11, 12, 25, 27, 30, 32, 34, 37, 40, 45, 47 \\
USB-A         & 11 & 18, 25, 26, 27, 28, 30, 44, 48, 50, 55, 58 \\
16\,mm cyl.   & 9  & 0, 17, 20, 22, 29, 30, 34, 37, 45 \\
HAN 10E       & 9  & 1, 5, 6, 8, 31, 34, 37, 38, 59 \\
NEMA 1-15P    & 8  & 15, 16, 17, 18, 19, 27, 51, 56 \\
RCA           & 7  & 8, 16, 22, 23, 25, 27, 31 \\
USB-C         & 1  & 17 \\
M12 fastener  & 0  & --- \\
\midrule
\textbf{Total} & \textbf{83} & \\
\bottomrule
\end{tabularx}
\end{table}

\subsection{Training on expert only demonstrations} 
\label{app:training}
All policy training was conducted on a single NVIDIA A100 GPU, running for approximately 24 hours per task. All models take as input the overhead, low, left-wrist, and right-wrist camera streams alongside 14-dimensional robot proprioception (joint positions and gripper states). Depth data is recorded and included as a dataset feature to support future work with depth-aware algorithms but is not used in any of the training runs. Each policy was rolled out for 15 episodes per task per configuration. Hyperparameters for each policy are reported below.

\noindent
\begin{minipage}[t]{0.32\textwidth}
\centering
\captionof{table}{ACT hyperparameters.}
\small
\begin{tabular}{@{}ll@{}}
\toprule
Learning rate     & 1e-5 \\
Batch size        & 16 \\
Encoder layers    & 1 \\
Optimizer              & AdamW \\
Decoder layers    & 4 \\
Feedforward dim.  & 3200 \\
Hidden dim.       & 512 \\
Attention heads   & 8 \\
Chunk size        & 100 \\
Dropout           & 0.1 \\
Vision backbone   & ResNet-18 \\
\bottomrule
\end{tabular}
\label{tab:act_hyperparams}
\end{minipage}%
\hfill
\begin{minipage}[t]{0.32\textwidth}
\centering
\captionof{table}{Diffusion Policy hyperparameters.}
\small
\begin{tabular}{@{}ll@{}}
\toprule
Learning rate          & 1e-4 \\
Batch size             & 64 \\
Optimizer              & AdamW \\
Weight decay           & 1e-6 \\
Observation steps      & 2 \\
Action steps           & 8 \\
Action horizon         & 16 \\
Drop last $n$ frames   & 7 \\
Diffusion steps        & 100 \\
Noise schedule         & DDPM \\
Hidden dim.            & 256 \\
Vision backbone        & ResNet-18 \\
\bottomrule
\end{tabular}
\label{tab:dp_hyperparams}
\end{minipage}%
\hfill
\begin{minipage}[t]{0.32\textwidth}
\centering
\captionof{table}{$\pi_0$-FAST fine-tuning hyperparameters.}
\small
\begin{tabular}{@{}ll@{}}
\toprule
Learning rate         & 2.5e-5 \\
PaliGemma variant     & gemma\_2b \\
Action expert variant & gemma\_300m \\
Optimizer             & AdamW \\
Chunk size            & 50 \\
Batch size            & -- \\
Vision encoder        & frozen \\
\bottomrule
\end{tabular}
\label{tab:pi0_hyperparams}
\end{minipage}

\subsection{Task inventory}
\label{app:task_inventory}

Table~\ref{tab:task_inventory} describes the 9 objects in REBOOT,
chosen to span a range of geometries, engagement tolerances, and
rotational symmetries while remaining representative of small-scale
assembly. Each object appears in both an install and a remove
variant, yielding 18 tasks in total. We report the rotational
symmetry group governing the alignment search space, the precision
tier based on engagement clearance, and the dominant failure modes
observed in autonomous policy rollouts.

\begin{table}[H]
\centering
\small
\caption{Task inventory for REBOOT. Each object appears as both an
install and a remove task. Symmetry: $C_1$ (one valid orientation),
$C_2$ (two valid orientations $180^\circ$ apart), Cont.\ (any
in-plane rotation valid). Tier: \emph{Int.}\ = interference fit,
\emph{Tight} = $<$0.2~mm, \emph{Mod.}\ = 0.2--1~mm, \emph{Loose} =
$>$1~mm.}
\label{tab:task_inventory}
\setlength{\tabcolsep}{5pt}
\renewcommand{\arraystretch}{1.15}
\begin{tabular}{l l c c c l}
\toprule
\textbf{Object} & \textbf{Mechanism} & \textbf{Sym.} &
\textbf{Tier} & \textbf{Clear.\ (mm)} &
\textbf{Object description} \\
\midrule
\multicolumn{6}{l}{\emph{\textbf{Threaded fastener}}} \\

M12 fastener & Threaded   & $C_2$    & Mod.  & 0.70 &
  Steel hex nut, 12~mm thread \\
\midrule
\multicolumn{6}{l}{\emph{\textbf{Electrical connectors}}} \\

USB-A         & Peg-in-hole & $C_1$    & Mod.  & 0.62 &
  Standard USB-A plug \\
USB-C         & Peg-in-hole & $C_2$    & Tight & 0.09 &
  Standard USB-C plug \\
RJ45      & Peg-in-hole     & $C_1$    & Tight & 0.13 &
  RJ45 plug with retention latch \\
NEMA 1-15P    & Peg-in-hole & $C_1$    & Mod.  & 0.50 &
  Two-prong AC power plug, polarized \\
HAN 10E       & Peg-in-hole      & $C_1$    & Loose & 2.73 &
  10-pin keyed industrial connector \\
RCA           & Peg-in-hole   & Cont.\   & Int.  & 0.00 &
  Coaxial audio plug \\
\midrule
\multicolumn{6}{l}{\emph{\textbf{Precision pegs}}} \\

16~mm bar     & Peg-in-hole & $C_2$    & Mod.  & 0.20 &
  Steel bar, 16~mm diameter \\
16~mm cyl.\   & Peg-in-hole & Cont.\   & Mod.  & 0.20 &
  Steel cylinder, 16~mm diameter \\
\bottomrule
\end{tabular}
\end{table}

\subsection{Social impact}
REBOOT supports research on robust manipulation policies for precision assembly. Positive impacts include progress toward robotic automation of tasks that are currently manual, repetitive, or ergonomically demanding; better recovery-aware policies could improve throughput and reduce worker injury in these settings. We view the dual-use risk as low given the focus on precision assembly, and the technology is not yet at a stage where workforce displacement is an immediate concern, though we acknowledge that broader adoption of recovery-aware manipulation in manufacturing will eventually require thoughtful workforce transition planning.



\end{document}